\documentclass{article} % For LaTeX2e
\usepackage{iclr2024_conference,times}
\iclrfinalcopy

\usepackage{amsmath,amsfonts,bm}

\def\eqref#1{equation~\ref{#1}}
\def\1{\bm{1}}

\DeclareMathAlphabet{\mathsfit}{\encodingdefault}{\sfdefault}{m}{sl}
\SetMathAlphabet{\mathsfit}{bold}{\encodingdefault}{\sfdefault}{bx}{n}

\usepackage{hyperref}
\usepackage{url}

\usepackage{amsfonts}       % blackboard math symbols
\usepackage{booktabs}       % professional-quality tables
\usepackage{graphicx}
\usepackage{wrapfig}
\usepackage{xcolor,colortbl}
\usepackage[normalem]{ulem}

\title{What Makes Pre-Trained Visual \\ Representations Successful \\ for Robust Manipulation?}

\author{
  Kaylee Burns$^{1}$ \quad Zach Witzel$^{1}$ \quad Jubayer Ibn Hamid$^1$ \quad Tianhe Yu$^{2}$ \\
  \textbf{Chelsea Finn}$^{1}$ \quad \textbf{Karol Hausman}$^{1,2}$\\
  $^1$Stanford University \qquad $^2$Google DeepMind
}

\begin{document}
\definecolor{manip}{RGB}{255, 168, 34}
\definecolor{sup}{RGB}{26, 192, 198} %#FF5733
\definecolor{selfsup}{RGB}{255, 97, 80}
\definecolor{other}{RGB}{128, 128, 128}
\newcommand{\kayl}[1]{\textcolor{red}{#1}}
\newcommand{\change}[1]{\textcolor{red}{#1}}

\maketitle

%===============================================================================

\begin{abstract}
Inspired by the success of transfer learning in computer vision, roboticists have investigated visual pre-training as a means to improve the learning efficiency and generalization ability of policies learned from pixels.
To that end, past work has favored large object interaction datasets, such as first-person videos of humans completing diverse tasks, in pursuit of manipulation-relevant features.
Although this approach improves the efficiency of policy learning, it remains unclear how reliable these representations are in the presence of distribution shifts that arise commonly in robotic applications.
Surprisingly, we find that visual representations designed for manipulation and control tasks do not necessarily generalize under subtle changes in lighting and scene texture or the introduction of distractor objects.
To understand what properties \textit{do} lead to robust representations, we compare the performance of 15 pre-trained vision models under different visual appearances.
We find that emergent segmentation ability is a strong predictor of out-of-distribution generalization among ViT models.
The rank order induced by this metric is more predictive than metrics that have previously guided generalization research within computer vision and machine learning, such as downstream ImageNet accuracy, in-domain accuracy, or shape-bias as evaluated by cue-conflict performance.
We test this finding extensively on a suite of distribution shifts in ten tasks across two simulated manipulation environments. On the ALOHA setup, segmentation score predicts real-world performance after offline training with 50 demonstrations.
Code and more information are available at: \url{https://kayburns.github.io/segmentingfeatures/}.
\end{abstract}


%===============================================================================

\section{Introduction}
\label{sec:intro}

In spite of vast progress in computer vision, the question of how to learn a good visual representation for robotics remains open~\citep{chen2021mocov3}. 
Elsewhere in computer vision, internet datasets are retrofit to new tasks with transfer learning, which promises both generalization and fast adaptation to downstream tasks in exchange for large-scale pre-training.
%
% For example, visual representations trained on ImageNet~\citep{deng2009imagenet} transfer successfully to tasks as diverse as medical image analysis~\citep{Morid2020ASR}, historical document analysis~\citep{Studer2019ACStudy}, and surface concrete crack detection~\citep{Dais2021automaticcrack}.
%
But in the field of robotics, this promise has yet to be fulfilled even though policies learned from pixels struggle substantially with data efficiency~\citep{Cobbe2018QuantifyingGI} and especially generalization under visual changes in a scene~\citep{Cobbe2019LeveragingPG}.

Recent work~\citep{Damen2018EPICKITCHENS,Grauman2022Ego4DAT} posits that the missing piece is a large pre-training dataset of object interactions across diverse environments --- the ImageNet~\citep{deng2009imagenet} or CommonCrawl~\citep{raffel2020exploringthelimits} of manipulation. That is, if we want to improve the visual generalization ability of pre-trained models we simply need to collect datasets of this kind at scale.
Indeed, training on large datasets of first-person human interaction data increases policy performance and learning efficiency downstream~\citep{nair2022r3m,Xiao2022mvp}, but these evaluations occur in environments that are very similar to those used for policy learning.
Robotic applications commonly contain environments with varying lighting conditions, scene textures, and background objects, and we want pre-trained representations to allow the robot to handle such variability.
Yet we have few concrete measures of how well pre-trained representations generalize out-of-distribution.
To take a step towards understanding these problems, our goal in this paper is to thoroughly answer the questions \textit{``which models generalize?''} and \textit{``how can we predict how well a pre-trained model will generalize?''}

\textbf{Our first key finding} is that, when evaluated under visual distribution shifts, models that are designed for manipulation and control do not outperform standard visual pre-training methods.
This finding violates our intuitions about what is needed to scale up robot learning and brings into question what constitutes relevant data, how to quantify useful features, and the importance of design choices such as model architecture. 
In other words, we need more guiding principles to help us understand what representations are good for manipulation and make the problem of iterating on pre-training strategies much more straightforward.
Currently, evaluating a pre-trained policy requires training and rolling out downstream policies across multiple environments and experimental conditions.
Instead, we can take inspiration from computer vision, which has developed proxies for robust performance on vast out-of-distribution datasets \citep{Geirhos2021PartialSI}.

\begin{wrapfigure}{R}{.5\textwidth}
\centering
% \vspace{-0.5cm}
    \includegraphics[width=.5\textwidth]{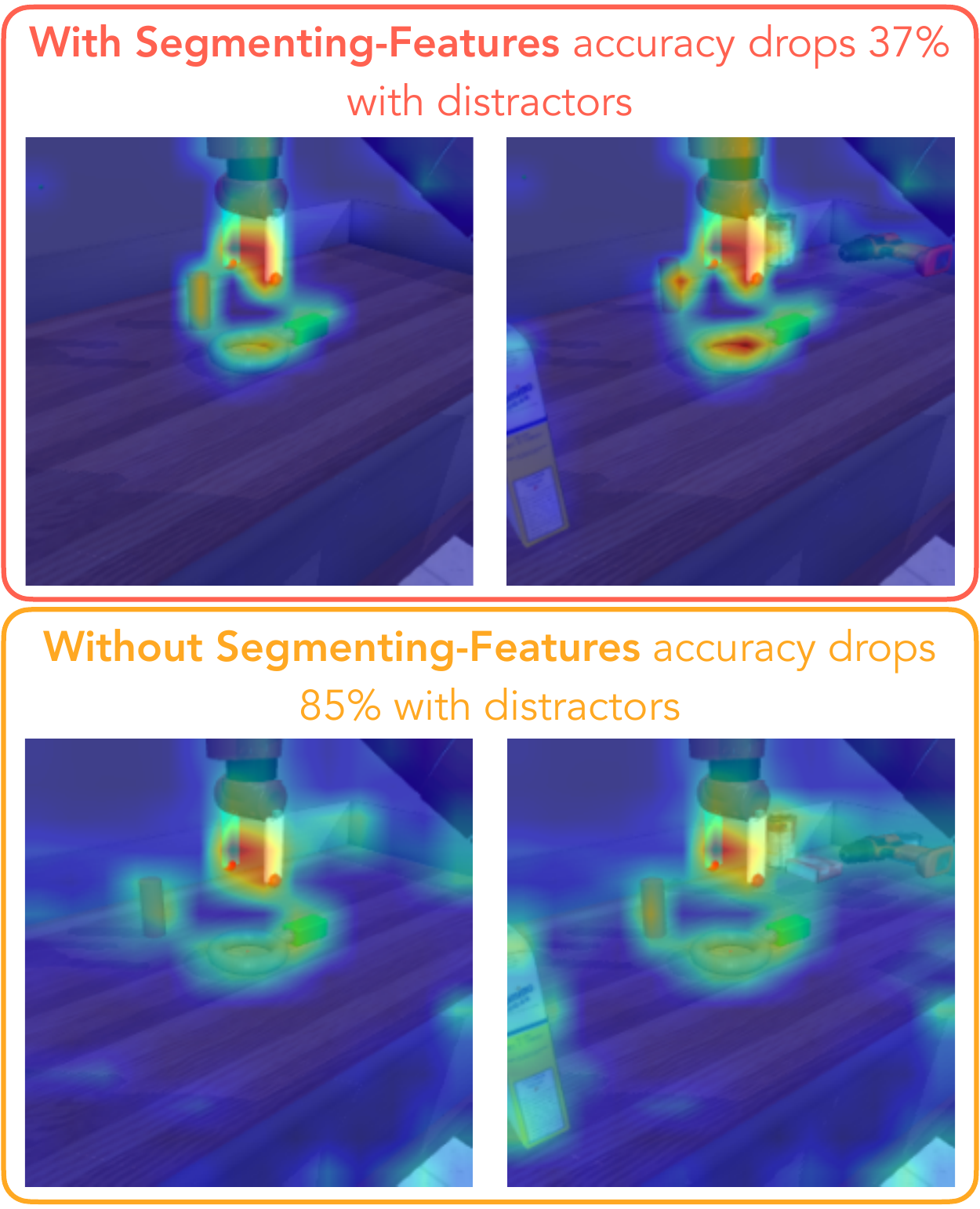}
    \caption{ We find that the emergent segmentation ability of ViT attention heads (measured by Jaccard index) predicts performance under visual distribution shift. We refer to models with this property as having ``segmenting-features.'' Notice how the attention of MVP shifts towards the sugar box distractor object in the bottom right image.
    The impact of this factor overshadows other design choices such as data relevance.}
    % \vspace{-10pt}
    \label{fig:heatmap_teaser}
\end{wrapfigure}

% First, we find that vision transformers (ViTs) generalize better than equivalently trained residual networks (ResNets). 
%
\textbf{Our second key finding} is that the emergent segmentation ability of a ViT model is a strong predictor of out-of-distribution generalization performance. We visualize this phenomenon, which we refer to as ``segmenting-features,'' in Figure~\ref{fig:heatmap_teaser}. 
Other metrics of model quality, such as linear probes on ImageNet~\citep{Chen2020ASF}, and metrics of out-of-distribution performance, such as in-domain accuracy~\citep{Miller2021AccuracyOT} and shape-bias~\citep{geirhos2018imagenettrained}, are not predictive for this model class, despite their predictive power in other commonly-studied domains like image classification. This hints at the possibility that the transfer setting of manipulation differs from computer vision tasks typically studied within the robustness literature. 

To reach the conclusions above, we run 9,000 different simulated evaluations. Our simulated environments are adapted from two different existing visual distribution shift benchmarks~\citep{xing2021kitchenshift,xie2023benchmarking} to capture the shifts that arise commonly in robotics applications: changes in lighting, background and object texture, and the appearance of distractors. More specifically, we train policies on top of 15 pre-trained models, including 4 models designed for manipulation or control: R3M~\citep{nair2022r3m}, two MVP variants~\citep{Xiao2022mvp,Radosavovic2022}, and VIP~\citep{ma2022vip}.
We further validate these findings by comparing a model designed for manipulation against a model with a similar parameter count on a real-world screwdriver pick-up task using the ACT training framework \citep{zhao2023LearningFG}. Through these experiments, we make two striking findings: (1) pre-trained visual models designed for control do not necessarily generalize better than models pre-trained on more standard computer vision datasets and (2) the emergent segmentation performance of a ViT model is a strong predictor of the out-of-distribution generalization of a down-stream policy.% 

% In addition to our experimental findings, we provide the complete set of environments and evaluation code to compare more pre-trained models at: \href{https://sites.google.com/stanford.edu/spatialfeatures}{https://sites.google.com/stanford.edu/spatialfeatures}.
% \vspace{-.15cm}

%===============================================================================

\section{Related Work}
\label{sec:rw}

\textbf{Representation learning for manipulation.}
The correct approach to visual representation learning for robotics is still an open question. 
There is evidence that separating visual representation learning from policy learning can further improve performance~\citep{Pari2022TheSE,Parisi2022TheUE}.
Recent works have shown that models pre-trained on large manipulation-relevant datasets~\citep{Goyal2017TheS,Damen2018EPICKITCHENS,Shan2020UnderstandingHH,Grauman2022Ego4DAT} or learned with visual affordances from RGBD data~\citep{lin2020learning} can improve the efficiency and performance of policy learning~\citep{karamcheti2023voltron} in comparison to standard vision datasets such as ImageNet~\citep{deng2009imagenet}, but they do not focus on performance under visual distribution shift. We evaluate the performance of R3M~\citep{nair2022r3m}, MVP~\citep{Xiao2022mvp,Radosavovic2022}, and VIP~\citep{ma2022vip}. Other work has studied generalization of pre-trained representations to new reinforcement learning tasks for manipulation~\citep{ma2022vip} and navigation~\citep{midLevelReps2018} where the agent is able to train on visual data from the new environment. 
Separate from the question of pre-training visual representations is the question of how to best train policies on top of pixel observations~\citep{laskin_srinivas2020curl,yarats2021image}.
\citet{majumdar2023vc1} benchmarks the performance of pre-trained visual representations on a handful of manipulation environments, but they focus on in-domain performance and also investigate navigation environments. \citet{hu2023pretrained} shows that model performance is highly sensitive to evaluation. We use imitation learning for our evaluation protocol, which they find to be a more stable measure of performance.
Concurrently with our work,~\cite{dasari2023datasets} demonstrates that the importance of proper data balancing supersedes the content of any one pre-training dataset. We focus on benchmarking visual generalization specifically and focus on advancing metrics that are predictive of generalization.

\textbf{Robustness in computer vision.}
There is extensive work studying the impact of design choices, such as architecture, loss, and data, on the performance of visual models under distribution shift. See~\citet{Geirhos2021PartialSI} for a comprehensive comparison. Most relevant to our paper are studies of shape-bias and architecture. While shape-biased models tend to be more robust than texture-biased ones~\citep{geirhos2018imagenettrained}, the impact of architecture on robustness is less straightforward. For example, vision transformers exhibit better robustness to universal adversarial attacks~\citep{shao2022on}, but they are more susceptible to patch-level attacks~\citep{fu2022patchfool}. When compared on natural distribution shifts~\citep{hendrycks2018benchmarking,Hendrycks2021manyfaces,hendrycks2021nae}, vision transformers and convolutional networks achieve comparable performance when provided with enough data~\citep{Bhojanapalli2021URtransformers}. But for occlusions specifically, vision transformers appear to have an edge~\citep{naseer2021intriguing}. \citet{Miller2021AccuracyOT} studies the predictive power of in-domain performance for out-of-distribution generalization. Unlike all of these prior works, we focus on how pre-trained representations affect robustness in downstream robotics tasks, instead of downstream vision tasks.

\begin{figure*}[t]
    \centering
    \includegraphics[width=\textwidth]{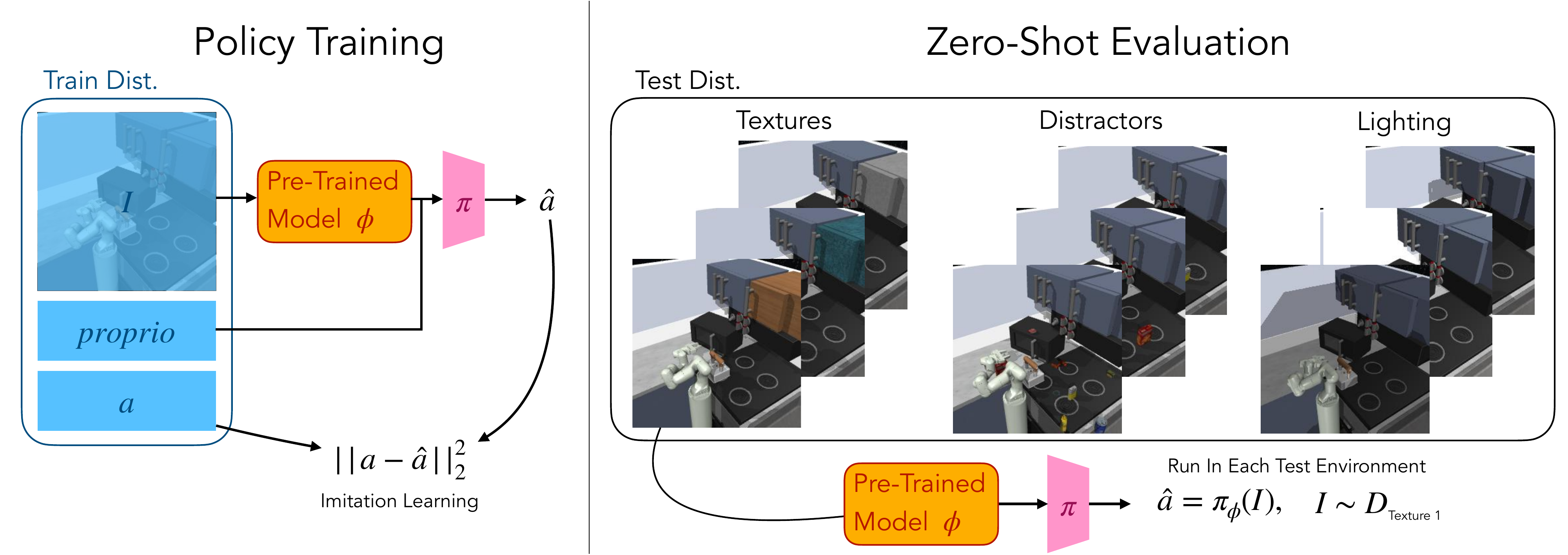}
    \caption{\textbf{Evaluation Scheme.} We begin our evaluation procedure by training a policy with behavior cloning on top of frozen features. In every experimental setting, we ablate the encoder used to extract features from the image observation. The learned policy is then evaluated in each of the visual shift environments to attain a zero-shot success value.}
    \label{fig:eval_setup}
    % \vspace{-10pt}
\end{figure*}

\textbf{Learning robust policies.}
Unlike work that focuses on changes in dynamics or initial state distribution~\citep{Huang2021AdaRLWW,Raileanu2020AutomaticDA,laskin2020reinforcement,cobbe2019procgen,Packer2018AssessingGI,Farebrother2018GeneralizationAR}, we focus exclusively on the setting of visual distribution shifts.
\citet{Kirk2021ASO} and~\citet{Zhao2019InvestigatingGI} provide a comprehensive survey on non-visual distribution shifts in decision making problems.
Policy adaptation approaches enable visual robustness specifically by leveraging insights from domain adaptation during policy training~\citep{hansen2021softda,fan2021secant,yoneda2021invariance} or during deployment~\citep{hansen2021deployment}.
In the special case of closing the sim-to-real domain gap, a popular approach is to add randomized textures while training in simulation~\citep{Sadeghi2017CAD2RLRS,Tobin2017DomainRF,Peng2018SimtoRealTO,James2019SimToRealVS}.
By contrast, our work is interested in explaining properties of a robust visual model for control. Consequently, our insights can be leveraged with or without any task specific data.
% 
	
%===============================================================================

\section{Environments, Evaluation Protocol, and Pre-Trained Models}

\label{sec:experimental_setup}
Our goal is to understand how robust existing representations for manipulation are to visual distribution shifts that are realistic in robotic applications.
To that end, we learn policies on top of frozen, pre-trained encoders and then evaluate these policies zero-shot under changes in lighting, object and scene texture, and the presence of distractors.
These shifts are visualized in Appendix Figure~\ref{fig:shift_vis} and a high level summary of our evaluation procedure is visualized in Figure~\ref{fig:eval_setup}.
In this section, we describe the specifics of the manipulation environments, distribution shifts, and policy training setups. 

\begin{figure*}[t]
    \centering
    \includegraphics[width=\textwidth]{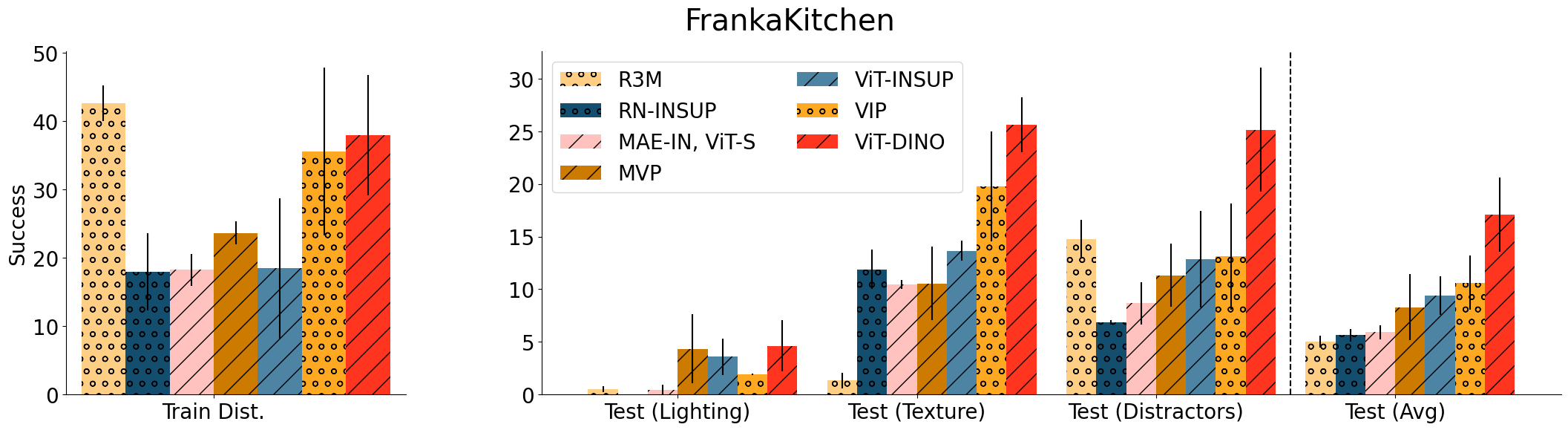}
    \includegraphics[width=\textwidth]{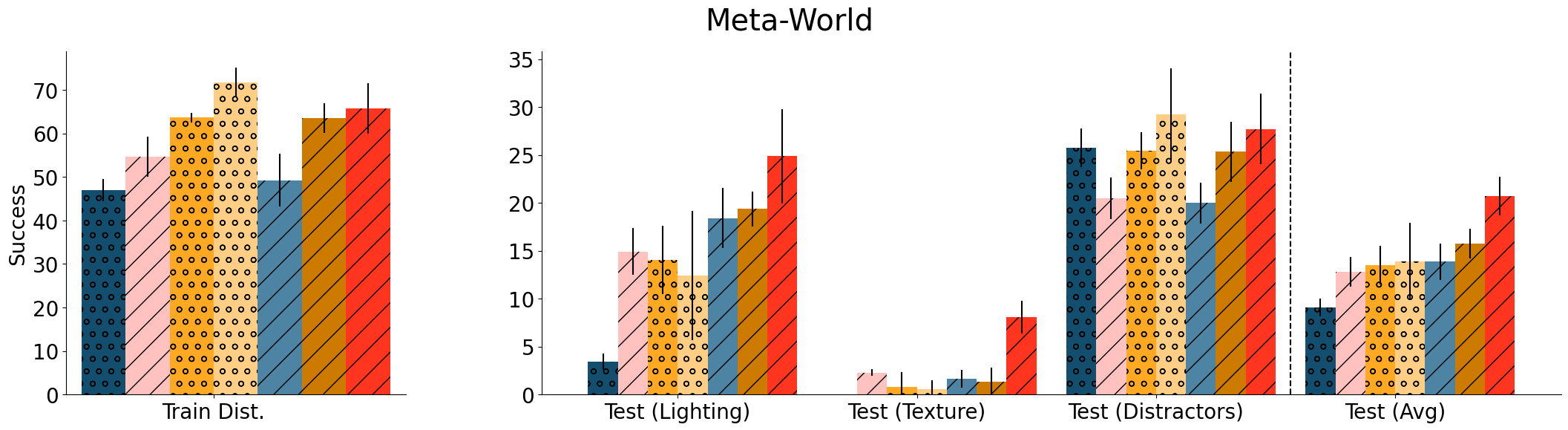}
    % \vspace{-0.6cm}
    \caption{\textbf{Visual Generalization Performance.} Models trained with supervision on ImageNet are shades of \textcolor{sup}{blue}. Models trained with self-supervision on ImageNet are in \textcolor{selfsup}{red}. Models trained explicitly for manipulation and control tasks are \textcolor{manip}{orange}. Dotted bars denote ResNets and slashed bars denote ViTs. Surprisingly, the best performing models are not necessarily the ones designed for manipulation. Each bar is an average over 30 experimental conditions.
    }
    % \vspace{-10pt}
    \label{fig:ood}
\end{figure*}

\textbf{Environments and tasks.} We study ten tasks across two simulated manipulation environments, which are selected based on their popularity in studying learning-based approaches to manipulation.
Within FrankaKitchen~\citep{gupta2020relay} we evaluate performance on opening a microwave, sliding a cabinet door open, pulling a cabinet open, turning a knob, and turning on a light. Within Meta-World~\citep{yu2019meta} we study assembling a ring onto a peg, placing an object between two bins, pushing a button, opening a drawer, and hammering a nail. 

\textbf{Distribution shifts.} We develop a benchmark for out-of-distribution generalization within FrankaKitchen and Meta-World. Within FrankaKitchen, we reimplement the texture and lighting changes from KitchenShift \citep{xing2021kitchenshift}. Within Meta-World we use texture changes from \citet{xie2023benchmarking} and reimplement the same lighting changes as in FrankaKitchen. In both environments we include three levels of distractors: one, three, and nine YCB objects \citep{calli2015ycb}.
More details about the implementation and parameterization of the distribution shifts are provided in Section~\ref{sec:dist_shift_details}.

\textbf{Policy training.} Policy training is done in the same manner as R3M \citep{nair2022r3m}. A summary of the evaluation scheme is provided in Figure~\ref{fig:eval_setup}. 
We train an MLP on top of the pre-trained embedding with imitation learning (IL), which, given actions sampled from expert trajectories, $a\sim \mathcal{D}_{{train}}$, minimizes the mean squared error objective, $||a - \hat{a}||_2^2$. Here $\hat{a}$ denotes the action predicted from a given policy. 
Details of the training procedure are provided in Section~\ref{sec:bc_details}.
The embedding weights are frozen during policy learning, so the pre-trained models receive no task data.
We train 3 different seeds within each task for each of two different camera angles. In total, we learn 60 policies for each model and perform 11 evaluations per policy, including on the train distribution.

Formally, for a pre-trained representation $\phi$ we learn policies, $\pi_\phi$, each trained with a different seed, camera angle, and task. We average the performance of $\pi_\phi$ along each experimental condition and compute the mean performance and error across seeds.

\textbf{Pre-trained Visual Representations.} We categorize pre-trained models by loss type and data source: supervised ImageNet models, self-supervised ImageNet models, and models trained for manipulation and control tasks. Model specifics are provided in Appendix Section~\ref{sec:model_details}.

\begin{wrapfigure}{R}{.5\textwidth}
\centering
  \includegraphics[width=.5\textwidth]{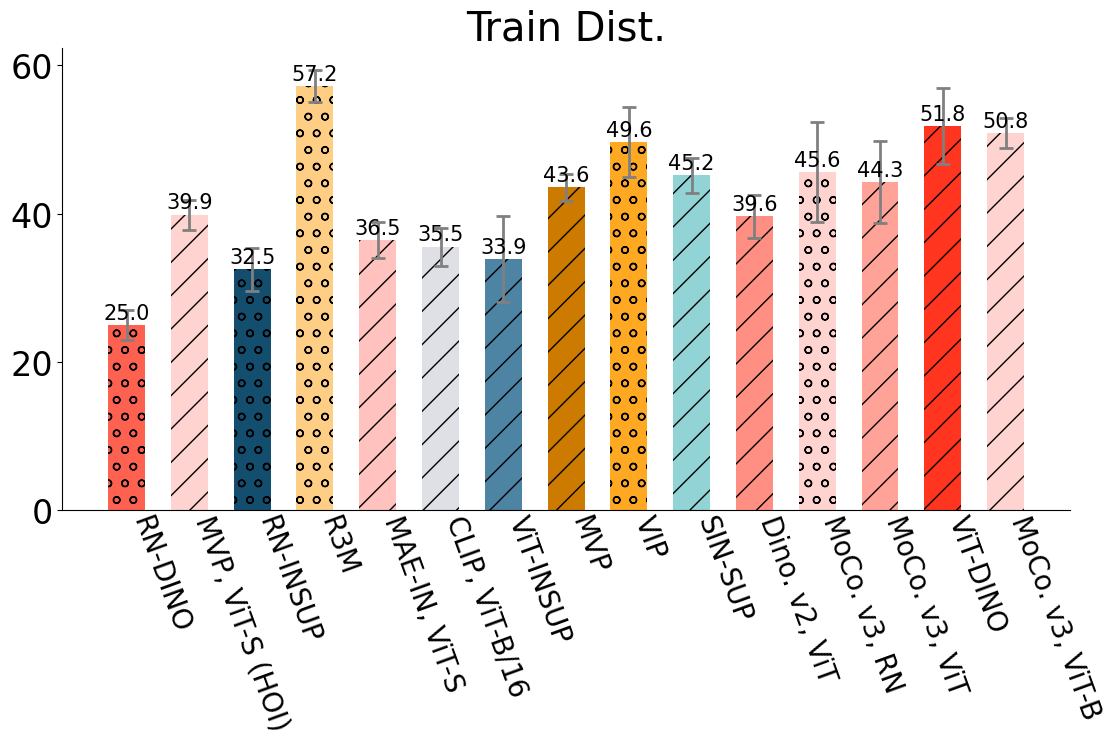}
  \includegraphics[width=.5\textwidth]{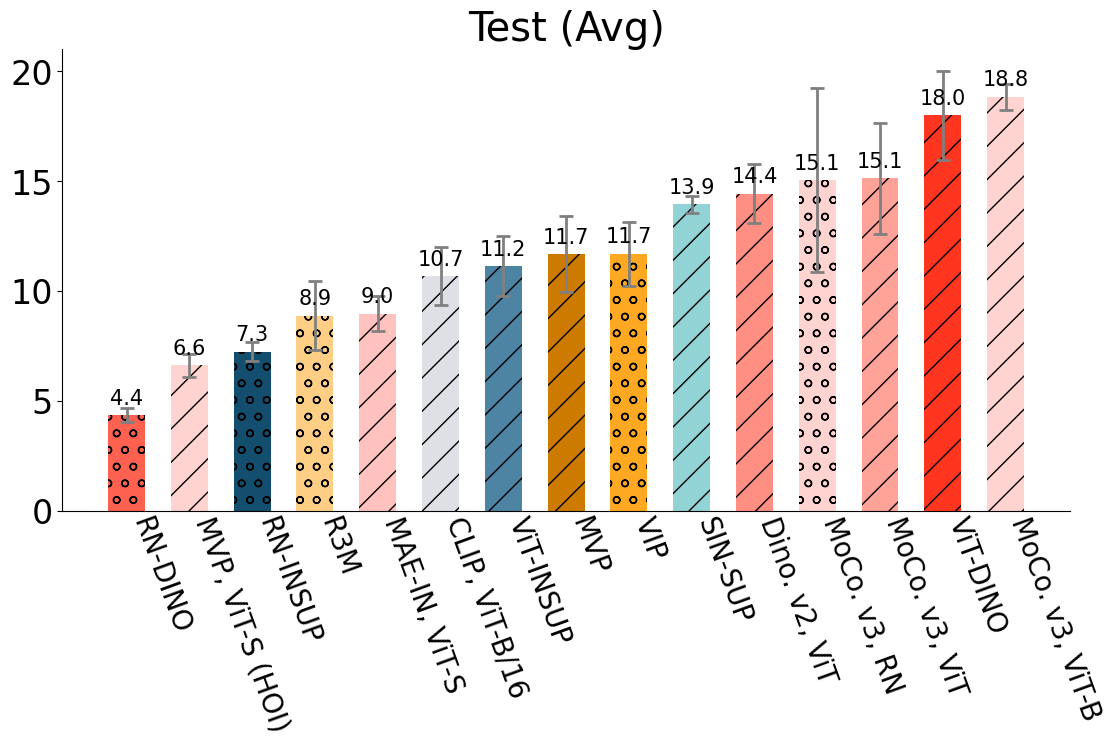}
    \caption{
    Average success rates for training and test distribution across both environments for every model in our evaluation suite. The best-performing model that was designed for manipulation ranks seventh out of all models evaluated.}
    % \vspace{-10pt}
    \label{fig:all_avg}
\end{wrapfigure}

\section{Generalization of Models Pre-Trained for Manipulation}
\label{sec:ood_performance}
% KH I don't think we need to motivate it based on scale, but just with the logic that the models pre-trained for manipulation should be better in general manipulation
% KB I brought out this point more in the introduction. lmk if it's still not cogent enough
One factor motivating work in learning-based robotics is the hypothesis of scale: if we collect more high-quality manipulation data, we should see improvements in policy generalization. 
However, our understanding of what high-quality data looks like for manipulation and control tasks is still imprecise. 
% 
% In this section, we test how well TODO(kayl): describe what will be compared and what the hypothesis is 
% 
Past work on pre-training visual representations for manipulation and control tasks has focused on collecting large object interaction datasets and developing manipulation-relevant losses. But the generalization ability of such models in comparison to standard pre-training methods is still unknown.
The goal of this section is to ask: \textit{which models generalize?}

To focus our analysis, we compare models pre-trained for manipulation to two self-supervised ImageNet models and two supervised ImageNet models.
Our main result is presented in Figure~\ref{fig:ood} where we plot the average success rate of the learned policies in the training environment distribution, within each class of visual shift, and across all types of visual shifts.

\begin{figure*}[t]
    \centering
    \includegraphics[width=\textwidth]{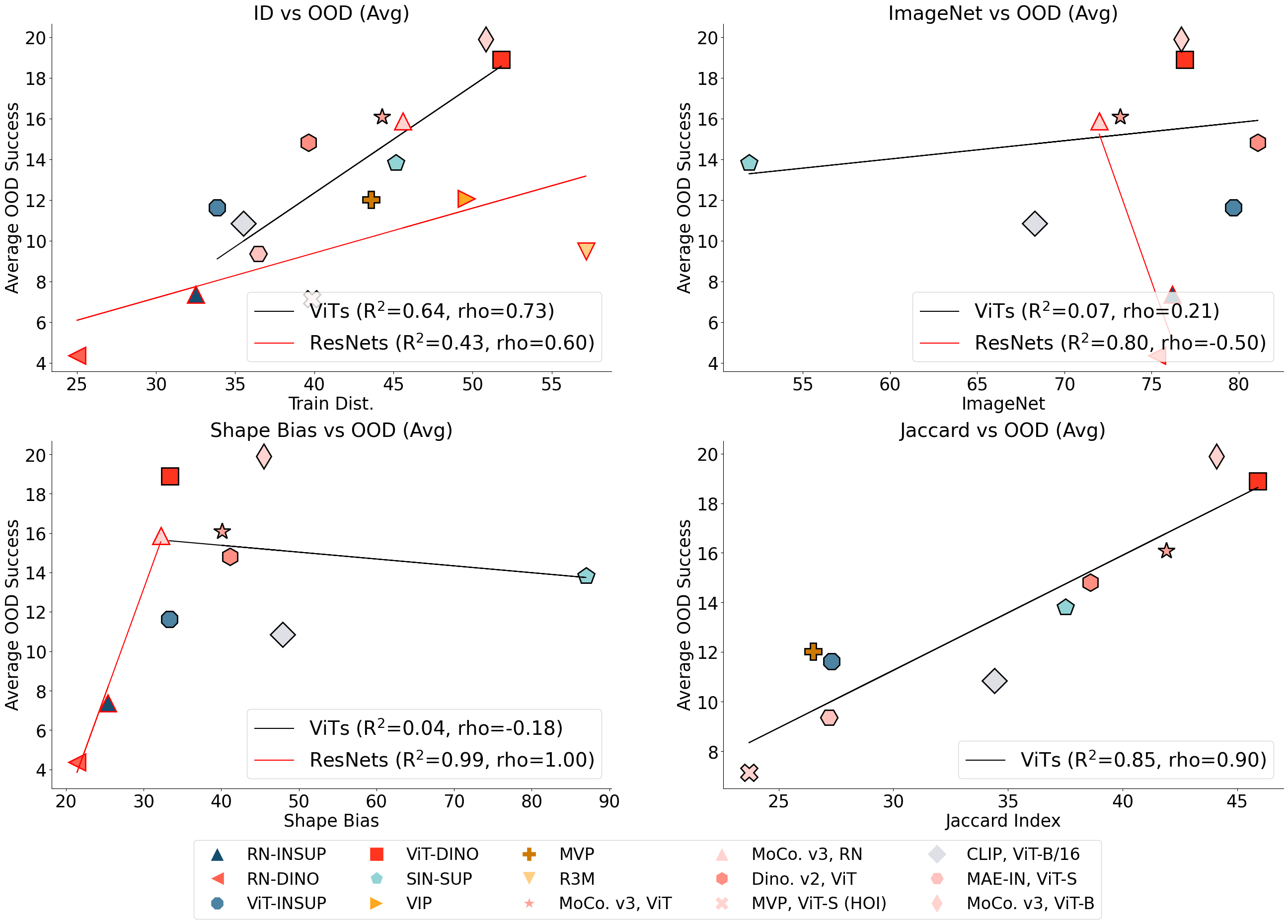}
    \caption{We plot the relationship between different metrics and out-of-distribution (OOD) generalization. There is a promising correlation between shape-bias and OOD performance for ResNets, but not ViTs. Instead, OOD performance for ViTs is strongly correlated with Jaccard index.}
    \vspace{-10pt}
    \label{fig:corr}
\end{figure*}

\textbf{\textcolor{manip}{Models pre-trained for manipulation.}} Past work has trained visual representations for manipulation in two ways: by training with manipulation-specific losses or on data of human-object interactions.
We focus on three recently introduced pre-trained models for manipulation that use different combinations of these approaches: Masked Visual Pretraining (MVP) \citep{Xiao2022mvp}, Reusable Representations for Robot Manipulation (R3M) \citep{nair2022r3m}, and Value-Implicit Pre-Training (VIP) \citep{ma2022vip}.
We include important characteristics of these models, including dataset sizes, architecture sizes, and augmentations in Section~\ref{sec:model_details} and Table~\ref{tab:pretrained_model_comparison}.

These models perform strongly within the training distribution: R3M and VIP in particular comfortably beat standard pre-training baselines. This is expected, especially for R3M which was evaluated on the same training environment. However, under subtle distribution shifts, models designed for manipulation struggle to generalize as well as supervised or self-supervised training with ImageNet. This is surprising for a few reasons. First, each manipulation model is trained on a larger dataset than the pre-trained baselines. Ego4D alone is 4.5M frames while ImageNet is only 1.2M. By parameter count, MVP is also larger than the ViT-S baselines. Finally, we expect human-object interaction datasets such as Ego4D to be more similar to the distribution of images observed when training a manipulation policy. The viewpoints are more varied and the scenes are less curated than ImageNet. Although we expect this to improve the generalization of the learned policy, these results show that other factors may supersede the impact of data relevance or scale alone. 

\textbf{\textcolor{sup}{Supervised ImageNet models.}} Supervised training on ImageNet has long been a baseline for visual pre-training. Past work has found that features learned with supervised learning on ImageNet are also a strong baseline for control: even frozen features are competitive with ground-truth state information on a variety of simulated control tasks \citep{Parisi2022TheUE}. However, \citet{Parisi2022TheUE} also find that self-supervised learning outperforms supervised learning. Our results contradict this finding. Figure~\ref{fig:all_avg} shows that supervised training on Stylized ImageNet achieves a higher success rate in the training distribution than self-supervised training on ImageNet with a masked auto-encoding loss.
These models maintain the same rank out-of-domain as well. Even without stylization, in-domain performance of supervised ImageNet models are competitive with models trained with MAE on FrankaKitchen. From these results, we conclude that the presence of supervision is not as predictive of in-domain or out-of-domain performance as other factors. We also find that supervised ImageNet training is still a strong baseline for model generalization: in both settings ViT-INSUP outperforms R3M and MVP.

\textbf{\textcolor{selfsup}{Self-Supervised ImageNet Models.}} In Figure~\ref{fig:ood} we include two self-supervised ViT-S models. Under visual distribution shifts, the model trained with the DINO objective outperforms all three models that are designed for manipulation. Moreover, this trend holds for every distribution shift except Meta-World with distractors. The distractors evaluation suite averages over different levels of distractions and therefore favors models with a high performance in training. In Appendix Section~\ref{sec:distractors} we plot model performance across different levels of distractors and find that several self-supervised ViTs experience a smaller drop in performance as more distractors are added compared to ResNet based pre-trained manipulation models like R3M and VIP.

Training with masked autoencoding performs well under distribution shifts in Meta-World, but is less strong under distribution shifts within FrankaKitchen. In Figure~\ref{fig:all_avg}, we see that MoCo. v3, ViT-B also performs strongly out-of-distribution. When we compare MoCo and DINO against MAE-style training we see that MoCo and DINO use a more extensive set of augmentations.
Taking this into account alongside the observation that a ViT trained with supervision on Stylized ImageNet performs well out-of-distribution we conclude that choice of augmentations outweighs the importance of supervision. This extends the findings of~\citet{Geirhos2021PartialSI} to the setting of robust manipulation. 

\begin{wrapfigure}{R}{.45\textwidth}
\centering
  \includegraphics[width=.45\textwidth]{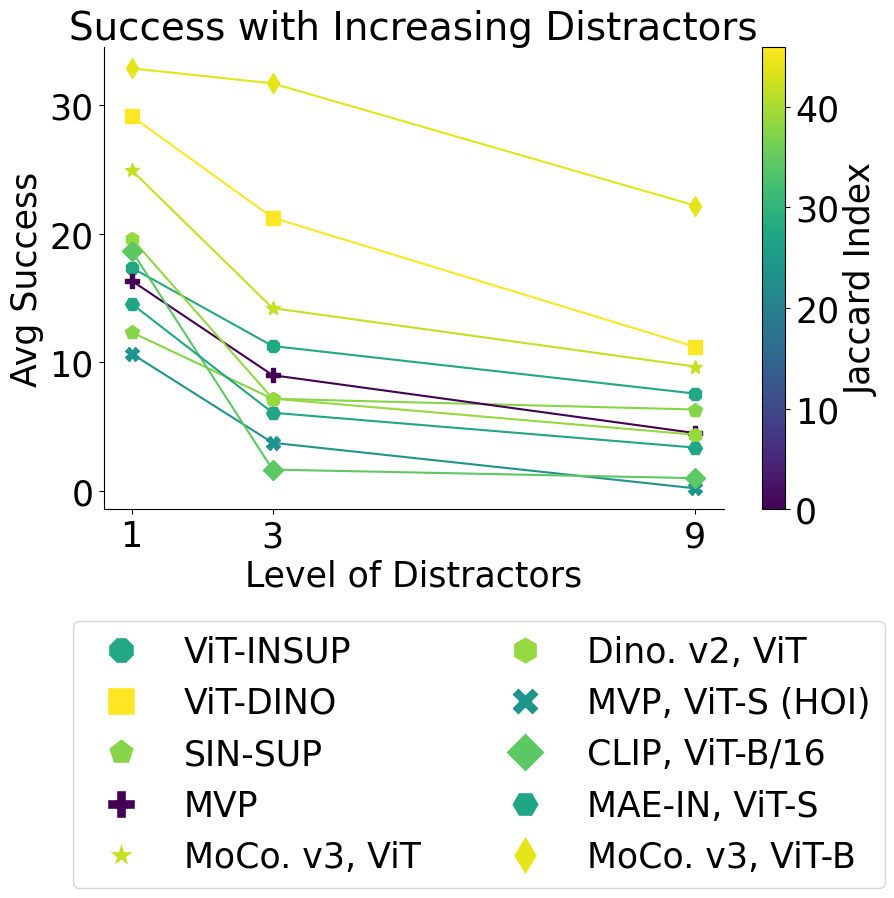}
    \caption{
    What happens to models with a high Jaccard index under an object-level distribution shift? Surprisingly, the models with the highest Jaccard index maintain the highest performance as the number of distractors increases.}
    % \vspace{-15pt}
    \label{fig:disrtactors_avg}
\end{wrapfigure}

\textbf{ViTs vs ResNets.}
One important design choice when selecting a pre-trained model is the choice of architecture. We focus on ResNets and ViTs.
% Convolutional networks are commonly used when learning control policies directly from pixels \citep{yarats2021image,Espeholt2018impala,Mnih2013PlayingAW}.
In all of our experiments, we use ResNet-50 \citep{He2016DeepRL} to be consistent with past work on visual pre-training \citep{Parisi2022TheUE,nair2022r3m,ma2022vip}. Vision transformers (ViTs) \citep{dosovitskiy2021an} have seen widespread adoption within computer vision \citep{khan2022TIVsurvey}, but have only recently been used for learning representations for control \citep{Xiao2022mvp}. We find that, on average, ViTs have a slight edge on out-of-distribution generalization compared to equivalently trained ResNets. In Figure~\ref{fig:disrtactors_avg}, out of the seven pre-trained models that perform best out-of-distribution six are ViTs. Ablating architecture alone while holding dataset, training augmentations, and parameter count constant, we can compare the model pairs ``MoCo. v3, RN'' and ``MoCo. v3, ViT'', ``RN-DINO'' and ``ViT-DINO'', and ``RN-INSUP'' and ``ViT-INSUP.'' In the latter two pairs, the ViT variant is much stronger out-of-distribution than the ResNet variant. For MoCo, the two variants achieve similar performance out-of-distribution.

\textbf{Summary.} This section identified which pre-trained models generalize, with several interesting findings. First, models designed for manipulaiton do not necessarily perform well under subtle distribution shifts in comparison to more standard pre-training methods. Second, the presence or absence of supervision does not matter as much as other factors on both in- and out-of-distribution generalization. Finally, ViTs have a slight edge over ResNets in out-of-distribution generalization.

\section{Properties of Robust Visual Representations for Manipulation}
\label{sec:properties}

Our findings in the last section are both surprising and somewhat unsatisfying because they contradict many of our intuitions about scale and generalization.
In our evaluation suite, we saw that better generalization is not cleanly explained by more data, bigger models, or more relevant data. 
The goal of this section is to identify the properties of pre-trained models that are predictive of generalization.
To that end, we correlate out-of-distribution performance with three metrics that have been previously connected to generalization in the machine learning and computer vision literature---in-domain performance, accuracy of a linear probe trained on ImageNet, and shape-bias. 
We also include a fourth metric, which is specific to ViTs: the emergent segmentation accuracy of the output attention heads. We describe each metric in detail in Section~\ref{sec:metrics}, discuss our setup for correlating performance in Section~\ref{sec:corr_setup}, and analyze our results in Section~\ref{sec:corr_results}.

\begin{wrapfigure}{R}{5.5cm}
\centering
\includegraphics[width=.3\textwidth]{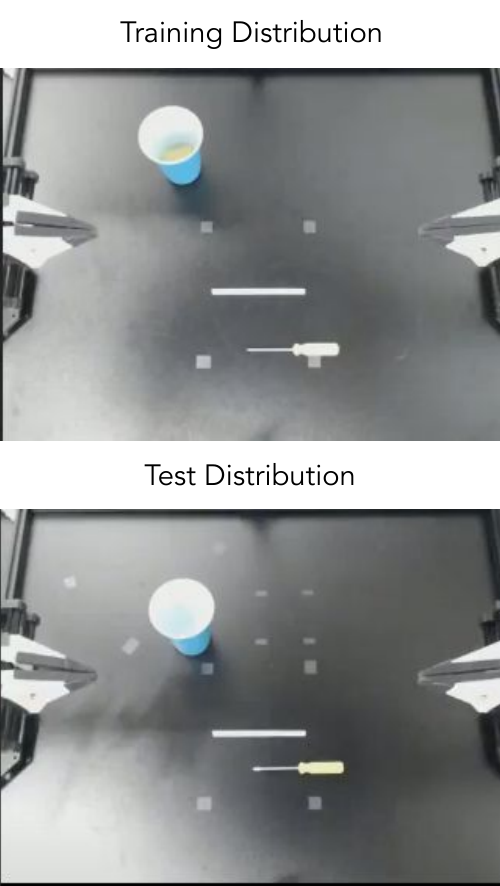}
    \caption{Real world training and test distribution. The test distribution differs from the training distribution in the position of the target objects and the direction of the lighting.
    }
    \label{fig:real_world}
\end{wrapfigure}

\subsection{Metrics}
\label{sec:metrics}
\textbf{ID vs OOD.}\label{sec:id_metric} One of the goals of this paper is to understand how well the findings from existing evaluations of pre-trained models hold under the inevitable environment changes that we expect to see in a real-world setting. If in-distribution performance is reasonably predictive of generalization to our suite of distribution shifts, it is sufficient for researchers to continue developing pre-trained models with existing methods of evaluation. Past work has also shown that the in-distribution performance of a pre-trained model is positively correlated with out-of-distribution performance for a variety of computer vision tasks \citep{Miller2021AccuracyOT}. Concretely, we measure in-distribution performance as the success rate of the policy within the training distribution.

\textbf{Imagenet vs OOD.} 
\label{sec:imagenet_metric}
Training linear probes on Imagenet is a common protocol for evaluating the quality of learned representations~\citep{He2019MomentumCF,Chen2020ASF}. \citet{hu2023pretrained} make the related finding that the ImageNet $k$-NN accuracy of a pre-trained model is predictive of performance on imitation learning with a visual reward function. We evaluate ImageNet validation set accuracy for all models with linear probes available.

\textbf{Shape-Bias vs OOD.} 
\label{sec:shapebias_metric}
Shape bias is the extent to which a model makes prediction decisions based on shape. We calculate shape bias as the percent of shape classification decisions out of the set of texture or shape classifications on the Stylized-ImageNet validation set \citep{geirhos2018imagenettrained} using the same probes described above.

\textbf{Jaccard vs OOD.} 
\label{sec:jaccard_metric}
Finally, for all of the ViT models, we look at the emergent segmentation performance. We evaluate the Jaccard index of an interpolated attention map averaged across heads in the last attention block at the [CLS] token.

\subsection{Setup}
\label{sec:corr_setup}
We measure the coefficient of determination ($R^2$) and Spearman's rank correlation ($\rho$) for the correlation between the out-of-distribution success rate and each metric described above. 
Our goal is to find a metric that will result in high correlation between the metric and the OOD success, i.e. both coefficients being close to $1.0$.
We fit separate trend lines to ViTs and ResNets. Because of the lack of available probes, we exclude MVP, MVP ViT-S HOI, R3M, VIP, and MAE-IN ViT-S from the shape bias and ImageNet probe correlations. Each point represents one of the 15 pre-trained models we evaluated and represents the average of 6,000 evaluation runs.

% \citet{Miller2021AccuracyOT} show that there is a linear trend between in-distribution and out-of-distribution performance for a large variety of learned models. However, some distribution shifts have weaker correlations. E.g., they find that specific image corruptions in CIFAR-10-C are more weakly correlated: Gaussian noise, Gaussian blur, shot noise, and speckle noise. They look at the amount of in-distribution data used and don't see a change in the trend. Formally in our experiments we look at in-distribution performance over the training distribution $D$ and compare it to the out-of-distribution performance $D'$. 

\subsection{Results}
\label{sec:corr_results}

We visualize the correlation between each metric and the average out-of-distribution success rate in Figure~\ref{fig:corr}. Although we see a positive relationship between in- and out-of distribution generalization, there are pre-trained models that notably deviate from this trend.
Among ViT models one example is MVP, ViT-S (HOI): the average success rate of this model drops to 6.63 from 39.86.
By contrast, we find that ImageNet accuracy of a linear probe poorly predicts generalization performance for ViTs.

We also see little correlation between shape-bias and OOD performance for ViT models, but a promisingly strong correlation on the subset of ResNets evaluted. This is surprising because humans make highly shape-biased decisions and increasing shape-bias increases the robustness of imagenet trained CNNs \citep{geirhos2018imagenettrained,Geirhos2021PartialSI}. One explanation of this finding is that the ViT architecture obviates the need for shape-biased features. For example, a ResNet-50 trained with the DINO training scheme has a strong shape-bias, but not the equivalent ViT model. 

Finally, we visualize the relationship between the Jaccard index and OOD performance on all ViT models in Figure~\ref{fig:corr}. There is a strong positive correlation between Jaccard index and OOD performance both in terms of rank correlation and the coefficient of determination. These results suggest that while shape-bias may not be predictive of the OOD generalization ability of a pre-trained ViT, the segmentation ability is a predictive alternative.

One counter-argument to the use of Jaccard index as a metric for for OOD performance is that it would be less predictive for object-level distribution shift, which would occur any time a large distractor is placed in the background of the image.
In Figure~\ref{fig:disrtactors_avg}, we plot the success rates of each ViT model as the number of objects increases and verify that the models with the higher Jaccard index actually maintain the highest performance as the number of distractors increases.
% Because the Jaccard index measures a proxy for the degree to which a model is object-centric, it's reasonable to question how models with a high Jaccard index perform under an object-level distribution shift.
% 
% In Figure~\ref{fig:disrtactors_avg}, we plot the success rates of each ViT model as the number of objects increases. 
% 
% Surpisingly, the models with the highest Jaccard index maintain the highest performance over the number of distractors.

\subsection{Validating in the real world}
In this section, we validate our finding on a real-world generalization scenario by comparing a ViT-B model designed for control (MVP) against a model not designed for control but with a high emergent segmentation score (MoCo-v3).

\begin{wraptable}{R}{5.5cm}
      \centering
    \begin{tabular}{cc}
    \toprule
    Model & Success \\
    \midrule
    MVP & 0\% \\
    MoCo-v3 & 40\% \\
    \end{tabular}%
  \caption{Success rates on the task of picking up the screwdriver.}
  \label{tab:real_success}%
\end{wraptable}

\textbf{Setup.} We learn policies for picking up a screwdriver on the ALOHA setup using the ACT training framework \citep{zhao2023LearningFG}. The training dataset is comprised of 50 episodes collected by an expert human demonstrator. Images are collected from 4 camera view points (one on each wrist, one top camera, and one front camera). We replace the standard encoder with a ViT-B and change the initialization of the encoder based on the experimental condition (i.e., we select for a different pre-trained model). We follow the standard ACT training paradigm with the hyperparameters listed in Appendix Table ~\ref{tab:hyperparameters_real}. From the training data to the test runs there is a distribution shift in both the placement of the target object (the screwdriver) and in the direction of the lighting. This is visualized in Figure~\ref{fig:real_world}. We calculate success on screw pick ups averaged over 10 rollouts in the test environment.

\textbf{Results.} We find that MoCo-v3 is stronger on this setting than MVP, even though it is not explicitly designed for manipulation. We find that the MoCo-v3 initialized encoder is able to achieve a success rate of 40\% on this task while the MVP initialized encoder is not able to successfully grasp the target object. Qualitatively, the MVP model fails in localizing the object when attempting the grasp, whereas MoCo-v3 model reliably localizes the object, but experiences more failure in finding the right grasp point.

% \vspace{-.25cm}
\section{Conclusion}
% \vspace{-.25cm}
 In this paper, we make several surprising findings about the generalization ability of pre-trained visual representations for manipulation tasks. First, we find that, contrary to the current direction in the literature, models pre-trained on manipulation-relevant data do no necessarily generalize better than models trained on standard pre-training datasets (such as ImageNet). Instead, we uncover a recipe for strong generalization: ViT models with a high emergent segmentation accuracy generalize well under visual distribution shifts. Emergent segmentation accuracy is not only a stronger predictor of generalization than many other metrics for robustness, but also requires no additional training to evaluate. This insight can guide the development of pre-trained vision models in future work: preferring architecture development and training algorithms that lead to strong emergent segmentation as opposed to only training on more manipulation-relevant data.

\subsubsection*{Acknowledgments}
This work was supported by ONR Grant N00014-22-1-2621. KB is supported by an NSF Fellowship. We thank Annie Xie, Kyle Hsu, Tony Zhao and Ananya Kumar for helpful feedback and support.

\bibliography{iclr2024_conference}

\begin{thebibliography}{71}
\providecommand{\natexlab}[1]{#1}
\providecommand{\url}[1]{\texttt{#1}}
\expandafter\ifx\csname urlstyle\endcsname\relax
  \providecommand{\doi}[1]{doi: #1}\else
  \providecommand{\doi}{doi: \begingroup \urlstyle{rm}\Url}\fi

\bibitem[Bhojanapalli et~al.(2021)Bhojanapalli, Chakrabarti, Glasner, Li,
  Unterthiner, and Veit]{Bhojanapalli2021URtransformers}
Srinadh Bhojanapalli, Ayan Chakrabarti, Daniel Glasner, Daliang Li, Thomas
  Unterthiner, and Andreas Veit.
\newblock Understanding robustness of transformers for image classification.
\newblock In \emph{2021 IEEE/CVF International Conference on Computer Vision
  (ICCV)}, pp.\  10211--10221, 2021.
\newblock \doi{10.1109/ICCV48922.2021.01007}.

\bibitem[Bommasani et~al.(2022)Bommasani, Hudson, Adeli, Altman, Arora, von
  Arx, Bernstein, Bohg, Bosselut, Brunskill, Brynjolfsson, Buch, Card,
  Castellon, Chatterji, Chen, Creel, Davis, Demszky, Donahue, Doumbouya,
  Durmus, Ermon, Etchemendy, Ethayarajh, Fei-Fei, Finn, Gale, Gillespie, Goel,
  Goodman, Grossman, Guha, Hashimoto, Henderson, Hewitt, Ho, Hong, Hsu, Huang,
  Icard, Jain, Jurafsky, Kalluri, Karamcheti, Keeling, Khani, Khattab, Koh,
  Krass, Krishna, Kuditipudi, Kumar, Ladhak, Lee, Lee, Leskovec, Levent, Li,
  Li, Ma, Malik, Manning, Mirchandani, Mitchell, Munyikwa, Nair, Narayan,
  Narayanan, Newman, Nie, Niebles, Nilforoshan, Nyarko, Ogut, Orr,
  Papadimitriou, Park, Piech, Portelance, Potts, Raghunathan, Reich, Ren, Rong,
  Roohani, Ruiz, Ryan, Ré, Sadigh, Sagawa, Santhanam, Shih, Srinivasan,
  Tamkin, Taori, Thomas, Tramèr, Wang, Wang, Wu, Wu, Wu, Xie, Yasunaga, You,
  Zaharia, Zhang, Zhang, Zhang, Zhang, Zheng, Zhou, and
  Liang]{bommasani2022opportunities}
Rishi Bommasani, Drew~A. Hudson, Ehsan Adeli, Russ Altman, Simran Arora, Sydney
  von Arx, Michael~S. Bernstein, Jeannette Bohg, Antoine Bosselut, Emma
  Brunskill, Erik Brynjolfsson, Shyamal Buch, Dallas Card, Rodrigo Castellon,
  Niladri Chatterji, Annie Chen, Kathleen Creel, Jared~Quincy Davis, Dora
  Demszky, Chris Donahue, Moussa Doumbouya, Esin Durmus, Stefano Ermon, John
  Etchemendy, Kawin Ethayarajh, Li~Fei-Fei, Chelsea Finn, Trevor Gale, Lauren
  Gillespie, Karan Goel, Noah Goodman, Shelby Grossman, Neel Guha, Tatsunori
  Hashimoto, Peter Henderson, John Hewitt, Daniel~E. Ho, Jenny Hong, Kyle Hsu,
  Jing Huang, Thomas Icard, Saahil Jain, Dan Jurafsky, Pratyusha Kalluri,
  Siddharth Karamcheti, Geoff Keeling, Fereshte Khani, Omar Khattab, Pang~Wei
  Koh, Mark Krass, Ranjay Krishna, Rohith Kuditipudi, Ananya Kumar, Faisal
  Ladhak, Mina Lee, Tony Lee, Jure Leskovec, Isabelle Levent, Xiang~Lisa Li,
  Xuechen Li, Tengyu Ma, Ali Malik, Christopher~D. Manning, Suvir Mirchandani,
  Eric Mitchell, Zanele Munyikwa, Suraj Nair, Avanika Narayan, Deepak
  Narayanan, Ben Newman, Allen Nie, Juan~Carlos Niebles, Hamed Nilforoshan,
  Julian Nyarko, Giray Ogut, Laurel Orr, Isabel Papadimitriou, Joon~Sung Park,
  Chris Piech, Eva Portelance, Christopher Potts, Aditi Raghunathan, Rob Reich,
  Hongyu Ren, Frieda Rong, Yusuf Roohani, Camilo Ruiz, Jack Ryan, Christopher
  Ré, Dorsa Sadigh, Shiori Sagawa, Keshav Santhanam, Andy Shih, Krishnan
  Srinivasan, Alex Tamkin, Rohan Taori, Armin~W. Thomas, Florian Tramèr,
  Rose~E. Wang, William Wang, Bohan Wu, Jiajun Wu, Yuhuai Wu, Sang~Michael Xie,
  Michihiro Yasunaga, Jiaxuan You, Matei Zaharia, Michael Zhang, Tianyi Zhang,
  Xikun Zhang, Yuhui Zhang, Lucia Zheng, Kaitlyn Zhou, and Percy Liang.
\newblock On the opportunities and risks of foundation models, 2022.

\bibitem[Calli et~al.(2015)Calli, Singh, Walsman, Srinivasa, Abbeel, and
  Dollar]{calli2015ycb}
Berk Calli, Arjun Singh, Aaron Walsman, Siddhartha Srinivasa, Pieter Abbeel,
  and Aaron~M. Dollar.
\newblock The ycb object and model set: Towards common benchmarks for
  manipulation research.
\newblock In \emph{2015 International Conference on Advanced Robotics (ICAR)},
  pp.\  510--517, 2015.
\newblock \doi{10.1109/ICAR.2015.7251504}.

\bibitem[Caron et~al.(2021)Caron, Touvron, Misra, J\'egou, Mairal, Bojanowski,
  and Joulin]{caron2021emerging}
Mathilde Caron, Hugo Touvron, Ishan Misra, Herv\'e J\'egou, Julien Mairal,
  Piotr Bojanowski, and Armand Joulin.
\newblock Emerging properties in self-supervised vision transformers.
\newblock In \emph{Proceedings of the International Conference on Computer
  Vision (ICCV)}, 2021.

\bibitem[Chen et~al.(2020)Chen, Kornblith, Norouzi, and Hinton]{Chen2020ASF}
Ting Chen, Simon Kornblith, Mohammad Norouzi, and Geoffrey~E. Hinton.
\newblock A simple framework for contrastive learning of visual
  representations.
\newblock \emph{ArXiv}, abs/2002.05709, 2020.

\bibitem[Chen* et~al.(2021)Chen*, Xie*, and He]{chen2021mocov3}
Xinlei Chen*, Saining Xie*, and Kaiming He.
\newblock An empirical study of training self-supervised vision transformers.
\newblock \emph{arXiv preprint arXiv:2104.02057}, 2021.

\bibitem[Cobbe et~al.(2018)Cobbe, Klimov, Hesse, Kim, and
  Schulman]{Cobbe2018QuantifyingGI}
Karl Cobbe, Oleg Klimov, Christopher Hesse, Taehoon Kim, and John Schulman.
\newblock Quantifying generalization in reinforcement learning.
\newblock \emph{ArXiv}, abs/1812.02341, 2018.

\bibitem[Cobbe et~al.(2019{\natexlab{a}})Cobbe, Hesse, Hilton, and
  Schulman]{Cobbe2019LeveragingPG}
Karl Cobbe, Christopher Hesse, Jacob Hilton, and John Schulman.
\newblock Leveraging procedural generation to benchmark reinforcement learning.
\newblock \emph{ArXiv}, abs/1912.01588, 2019{\natexlab{a}}.

\bibitem[Cobbe et~al.(2019{\natexlab{b}})Cobbe, Hesse, Hilton, and
  Schulman]{cobbe2019procgen}
Karl Cobbe, Christopher Hesse, Jacob Hilton, and John Schulman.
\newblock Leveraging procedural generation to benchmark reinforcement learning.
\newblock \emph{arXiv preprint arXiv:1912.01588}, 2019{\natexlab{b}}.

\bibitem[Damen et~al.(2018)Damen, Doughty, Farinella, Fidler, Furnari, Kazakos,
  Moltisanti, Munro, Perrett, Price, and Wray]{Damen2018EPICKITCHENS}
Dima Damen, Hazel Doughty, Giovanni~Maria Farinella, Sanja Fidler, Antonino
  Furnari, Evangelos Kazakos, Davide Moltisanti, Jonathan Munro, Toby Perrett,
  Will Price, and Michael Wray.
\newblock Scaling egocentric vision: The epic-kitchens dataset.
\newblock In \emph{European Conference on Computer Vision (ECCV)}, 2018.

\bibitem[Dasari et~al.(2023)Dasari, Srirama, Jain, and
  Gupta]{dasari2023datasets}
Sudeep Dasari, Mohan~Kumar Srirama, Unnat Jain, and Abhinav Gupta.
\newblock An unbiased look at datasets for visuo-motor pre-training.
\newblock In \emph{Conference on Robot Learning}. PMLR, 2023.

\bibitem[Deng et~al.(2009)Deng, Dong, Socher, Li, Li, and
  Fei-Fei]{deng2009imagenet}
Jia Deng, Wei Dong, Richard Socher, Li-Jia Li, Kai Li, and Li~Fei-Fei.
\newblock Imagenet: A large-scale hierarchical image database.
\newblock In \emph{2009 IEEE Conference on Computer Vision and Pattern
  Recognition}, pp.\  248--255, 2009.
\newblock \doi{10.1109/CVPR.2009.5206848}.

\bibitem[Dosovitskiy et~al.(2021)Dosovitskiy, Beyer, Kolesnikov, Weissenborn,
  Zhai, Unterthiner, Dehghani, Minderer, Heigold, Gelly, Uszkoreit, and
  Houlsby]{dosovitskiy2021an}
Alexey Dosovitskiy, Lucas Beyer, Alexander Kolesnikov, Dirk Weissenborn,
  Xiaohua Zhai, Thomas Unterthiner, Mostafa Dehghani, Matthias Minderer, Georg
  Heigold, Sylvain Gelly, Jakob Uszkoreit, and Neil Houlsby.
\newblock An image is worth 16x16 words: Transformers for image recognition at
  scale.
\newblock In \emph{International Conference on Learning Representations}, 2021.
\newblock URL \url{https://openreview.net/forum?id=YicbFdNTTy}.

\bibitem[Fan et~al.(2021)Fan, Wang, Huang, Yu, Fei-Fei, Zhu, and
  Anandkumar]{fan2021secant}
Linxi Fan, Guanzhi Wang, De-An Huang, Zhiding Yu, Li~Fei-Fei, Yuke Zhu, and
  Animashree Anandkumar.
\newblock Secant: Self-expert cloning for zero-shot generalization of visual
  policies.
\newblock In Marina Meila and Tong Zhang (eds.), \emph{Proceedings of the 38th
  International Conference on Machine Learning}, volume 139 of
  \emph{Proceedings of Machine Learning Research}, pp.\  3088--3099. PMLR,
  18--24 Jul 2021.
\newblock URL \url{https://proceedings.mlr.press/v139/fan21c.html}.

\bibitem[Farebrother et~al.(2018)Farebrother, Machado, and
  Bowling]{Farebrother2018GeneralizationAR}
Jesse Farebrother, Marlos~C. Machado, and Michael~H. Bowling.
\newblock Generalization and regularization in dqn.
\newblock \emph{ArXiv}, abs/1810.00123, 2018.

\bibitem[Fu et~al.(2022)Fu, Zhang, Wu, Wan, and Lin]{fu2022patchfool}
Yonggan Fu, Shunyao Zhang, Shang Wu, Cheng Wan, and Yingyan Lin.
\newblock Patch-fool: Are vision transformers always robust against adversarial
  perturbations?
\newblock In \emph{International Conference on Learning Representations}, 2022.
\newblock URL \url{https://openreview.net/forum?id=28ib9tf6zhr}.

\bibitem[Geirhos et~al.(2019)Geirhos, Rubisch, Michaelis, Bethge, Wichmann, and
  Brendel]{geirhos2018imagenettrained}
Robert Geirhos, Patricia Rubisch, Claudio Michaelis, Matthias Bethge, Felix~A.
  Wichmann, and Wieland Brendel.
\newblock Imagenet-trained {CNN}s are biased towards texture; increasing shape
  bias improves accuracy and robustness.
\newblock In \emph{International Conference on Learning Representations}, 2019.
\newblock URL \url{https://openreview.net/forum?id=Bygh9j09KX}.

\bibitem[Geirhos et~al.(2021)Geirhos, Narayanappa, Mitzkus, Thieringer, Bethge,
  Wichmann, and Brendel]{Geirhos2021PartialSI}
Robert Geirhos, Kantharaju Narayanappa, Benjamin Mitzkus, Tizian Thieringer,
  Matthias Bethge, Felix~A. Wichmann, and Wieland Brendel.
\newblock Partial success in closing the gap between human and machine vision.
\newblock In \emph{Neural Information Processing Systems}, 2021.

\bibitem[Goyal et~al.(2017)Goyal, Kahou, Michalski, Materzynska, Westphal, Kim,
  Haenel, Fr{\"u}nd, Yianilos, Mueller-Freitag, Hoppe, Thurau, Bax, and
  Memisevic]{Goyal2017TheS}
Raghav Goyal, Samira~Ebrahimi Kahou, Vincent Michalski, Joanna Materzynska,
  Susanne Westphal, Heuna Kim, Valentin Haenel, Ingo Fr{\"u}nd, Peter~N.
  Yianilos, Moritz Mueller-Freitag, Florian Hoppe, Christian Thurau, Ingo Bax,
  and Roland Memisevic.
\newblock The “something something” video database for learning and
  evaluating visual common sense.
\newblock \emph{2017 IEEE International Conference on Computer Vision (ICCV)},
  pp.\  5843--5851, 2017.

\bibitem[Grauman et~al.(2022)Grauman, Westbury, Byrne, Chavis, Furnari,
  Girdhar, Hamburger, Jiang, Liu, Liu, Martin, Nagarajan, Radosavovic,
  Ramakrishnan, Ryan, Sharma, Wray, Xu, Xu, Zhao, Bansal, Batra, Cartillier,
  Crane, Do, Doulaty, Erapalli, Feichtenhofer, Fragomeni, Fu, Fuegen,
  Gebreselasie, Gonz{\'a}lez, Hillis, Huang, Huang, Jia, Khoo, Kol{\'a}r,
  Kottur, Kumar, Landini, Li, Li, Li, Mangalam, Modhugu, Munro, Murrell,
  Nishiyasu, Price, Puentes, Ramazanova, Sari, Somasundaram, Southerland,
  Sugano, Tao, Vo, Wang, Wu, Yagi, Zhu, Arbel{\'a}ez, Crandall, Damen,
  Farinella, Ghanem, Ithapu, Jawahar, Joo, Kitani, Li, Newcombe, Oliva, Park,
  Rehg, Sato, Shi, Shou, Torralba, Torresani, Yan, and
  Malik]{Grauman2022Ego4DAT}
Kristen Grauman, Andrew Westbury, Eugene Byrne, Zachary~Q. Chavis, Antonino
  Furnari, Rohit Girdhar, Jackson Hamburger, Hao Jiang, Miao Liu, Xingyu Liu,
  Miguel Martin, Tushar Nagarajan, Ilija Radosavovic, Santhosh~K. Ramakrishnan,
  Fiona Ryan, Jayant Sharma, Michael Wray, Mengmeng Xu, Eric~Z. Xu, Chen Zhao,
  Siddhant Bansal, Dhruv Batra, Vincent Cartillier, Sean Crane, Tien Do, Morrie
  Doulaty, Akshay Erapalli, Christoph Feichtenhofer, Adriano Fragomeni, Qichen
  Fu, Christian Fuegen, Abrham Gebreselasie, Cristina Gonz{\'a}lez, James~M.
  Hillis, Xuhua Huang, Yifei Huang, Wenqi Jia, Weslie Yu~Heng Khoo, J{\'a}chym
  Kol{\'a}r, Satwik Kottur, Anurag Kumar, Federico Landini, Chao Li, Yanghao
  Li, Zhenqiang Li, Karttikeya Mangalam, Raghava Modhugu, Jonathan Munro,
  Tullie Murrell, Takumi Nishiyasu, Will Price, Paola~Ruiz Puentes, Merey
  Ramazanova, Leda Sari, Kiran~K. Somasundaram, Audrey Southerland, Yusuke
  Sugano, Ruijie Tao, Minh Vo, Yuchen Wang, Xindi Wu, Takuma Yagi, Yunyi Zhu,
  Pablo Arbel{\'a}ez, David~J. Crandall, Dima Damen, Giovanni~Maria Farinella,
  Bernard Ghanem, Vamsi~Krishna Ithapu, C.~V. Jawahar, Hanbyul Joo, Kris
  Kitani, Haizhou Li, Richard~A. Newcombe, Aude Oliva, Hyun~Soo Park, James~M.
  Rehg, Yoichi Sato, Jianbo Shi, Mike~Zheng Shou, Antonio Torralba, Lorenzo
  Torresani, Mingfei Yan, and Jitendra Malik.
\newblock Ego4d: Around the world in 3,000 hours of egocentric video.
\newblock \emph{2022 IEEE/CVF Conference on Computer Vision and Pattern
  Recognition (CVPR)}, pp.\  18973--18990, 2022.

\bibitem[Gupta et~al.(2019)Gupta, Kumar, Lynch, Levine, and
  Hausman]{gupta2019relay}
Abhishek Gupta, Vikash Kumar, Corey Lynch, Sergey Levine, and Karol Hausman.
\newblock Relay policy learning: Solving long-horizon tasks via imitation and
  reinforcement learning.
\newblock \emph{arXiv preprint arXiv:1910.11956}, 2019.

\bibitem[Gupta et~al.(2020)Gupta, Kumar, Lynch, Levine, and
  Hausman]{gupta2020relay}
Abhishek Gupta, Vikash Kumar, Corey Lynch, Sergey Levine, and Karol Hausman.
\newblock Relay policy learning: Solving long-horizon tasks via imitation and
  reinforcement learning.
\newblock In \emph{Conference on Robot Learning}, pp.\  1025--1037. PMLR, 2020.

\bibitem[Hansen \& Wang(2021)Hansen and Wang]{hansen2021softda}
Nicklas Hansen and Xiaolong Wang.
\newblock Generalization in reinforcement learning by soft data augmentation.
\newblock In \emph{International Conference on Robotics and Automation}, 2021.

\bibitem[Hansen et~al.(2021)Hansen, Jangir, Sun, Alenyà, Abbeel, Efros, Pinto,
  and Wang]{hansen2021deployment}
Nicklas Hansen, Rishabh Jangir, Yu~Sun, Guillem Alenyà, Pieter Abbeel,
  Alexei~A. Efros, Lerrel Pinto, and Xiaolong Wang.
\newblock Self-supervised policy adaptation during deployment.
\newblock In \emph{International Conference on Learning Representations}, 2021.

\bibitem[He et~al.(2016)He, Zhang, Ren, and Sun]{He2016DeepRL}
Kaiming He, X.~Zhang, Shaoqing Ren, and Jian Sun.
\newblock Deep residual learning for image recognition.
\newblock \emph{2016 IEEE Conference on Computer Vision and Pattern Recognition
  (CVPR)}, pp.\  770--778, 2016.

\bibitem[He et~al.(2019)He, Fan, Wu, Xie, and Girshick]{He2019MomentumCF}
Kaiming He, Haoqi Fan, Yuxin Wu, Saining Xie, and Ross~B. Girshick.
\newblock Momentum contrast for unsupervised visual representation learning.
\newblock \emph{2020 IEEE/CVF Conference on Computer Vision and Pattern
  Recognition (CVPR)}, pp.\  9726--9735, 2019.

\bibitem[He et~al.(2021)He, Chen, Xie, Li, Doll'ar, and
  Girshick]{He2021MaskedAA}
Kaiming He, Xinlei Chen, Saining Xie, Yanghao Li, Piotr Doll'ar, and Ross~B.
  Girshick.
\newblock Masked autoencoders are scalable vision learners.
\newblock \emph{2022 IEEE/CVF Conference on Computer Vision and Pattern
  Recognition (CVPR)}, pp.\  15979--15988, 2021.

\bibitem[Hendrycks \& Dietterich(2019)Hendrycks and
  Dietterich]{hendrycks2018benchmarking}
Dan Hendrycks and Thomas Dietterich.
\newblock Benchmarking neural network robustness to common corruptions and
  perturbations.
\newblock In \emph{International Conference on Learning Representations}, 2019.
\newblock URL \url{https://openreview.net/forum?id=HJz6tiCqYm}.

\bibitem[Hendrycks et~al.(2021{\natexlab{a}})Hendrycks, Basart, Mu, Kadavath,
  Wang, Dorundo, Desai, Zhu, Parajuli, Guo, Song, Steinhardt, and
  Gilmer]{Hendrycks2021manyfaces}
Dan Hendrycks, Steven Basart, Norman Mu, Saurav Kadavath, Frank Wang, Evan
  Dorundo, Rahul Desai, Tyler Zhu, Samyak Parajuli, Mike Guo, Dawn Song, Jacob
  Steinhardt, and Justin Gilmer.
\newblock The many faces of robustness: A critical analysis of
  out-of-distribution generalization.
\newblock In \emph{ICCV}, pp.\  8320--8329, 2021{\natexlab{a}}.
\newblock URL \url{https://doi.org/10.1109/ICCV48922.2021.00823}.

\bibitem[Hendrycks et~al.(2021{\natexlab{b}})Hendrycks, Zhao, Basart,
  Steinhardt, and Song]{hendrycks2021nae}
Dan Hendrycks, Kevin Zhao, Steven Basart, Jacob Steinhardt, and Dawn Song.
\newblock Natural adversarial examples.
\newblock \emph{CVPR}, 2021{\natexlab{b}}.

\bibitem[Hu et~al.(2023)Hu, Wang, Li, and Gao]{hu2023pretrained}
Yingdong Hu, Renhao Wang, Li~Erran Li, and Yang Gao.
\newblock For pre-trained vision models in motor control, not all policy
  learning methods are created equal, 2023.

\bibitem[Huang et~al.(2021)Huang, Feng, Lu, Magliacane, and
  Zhang]{Huang2021AdaRLWW}
Biwei Huang, Fan Feng, Chaochao Lu, Sara Magliacane, and Kun Zhang.
\newblock Adarl: What, where, and how to adapt in transfer reinforcement
  learning.
\newblock \emph{ArXiv}, abs/2107.02729, 2021.

\bibitem[James et~al.(2019)James, Wohlhart, Kalakrishnan, Kalashnikov, Irpan,
  Ibarz, Levine, Hadsell, and Bousmalis]{James2019SimToRealVS}
Stephen James, Paul Wohlhart, Mrinal Kalakrishnan, Dmitry Kalashnikov, Alex
  Irpan, Julian Ibarz, Sergey Levine, Raia Hadsell, and Konstantinos Bousmalis.
\newblock Sim-to-real via sim-to-sim: Data-efficient robotic grasping via
  randomized-to-canonical adaptation networks.
\newblock \emph{2019 IEEE/CVF Conference on Computer Vision and Pattern
  Recognition (CVPR)}, pp.\  12619--12629, 2019.

\bibitem[Karamcheti et~al.(2023)Karamcheti, Nair, Chen, Kollar, Finn, Sadigh,
  and Liang]{karamcheti2023voltron}
Siddharth Karamcheti, Suraj Nair, Annie~S. Chen, Thomas Kollar, Chelsea Finn,
  Dorsa Sadigh, and Percy Liang.
\newblock Language-driven representation learning for robotics.
\newblock \emph{arXiv preprint arXiv:2302.12766}, 2023.

\bibitem[Khan et~al.(2022)Khan, Naseer, Hayat, Zamir, Khan, and
  Shah]{khan2022TIVsurvey}
Salman Khan, Muzammal Naseer, Munawar Hayat, Syed~Waqas Zamir, Fahad~Shahbaz
  Khan, and Mubarak Shah.
\newblock Transformers in vision: A survey.
\newblock \emph{ACM Comput. Surv.}, 54\penalty0 (10s), sep 2022.
\newblock ISSN 0360-0300.
\newblock \doi{10.1145/3505244}.
\newblock URL \url{https://doi.org/10.1145/3505244}.

\bibitem[Kirk et~al.(2021)Kirk, Zhang, Grefenstette, and
  Rocktaschel]{Kirk2021ASO}
Robert Kirk, Amy Zhang, Edward Grefenstette, and Tim Rocktaschel.
\newblock A survey of generalisation in deep reinforcement learning.
\newblock \emph{ArXiv}, abs/2111.09794, 2021.

\bibitem[Laskin et~al.(2020{\natexlab{a}})Laskin, Lee, Stooke, Pinto, Abbeel,
  and Srinivas]{laskin2020reinforcement}
Michael Laskin, Kimin Lee, Adam Stooke, Lerrel Pinto, Pieter Abbeel, and
  Aravind Srinivas.
\newblock Reinforcement learning with augmented data.
\newblock \emph{arXiv preprint arXiv:2004.14990}, 2020{\natexlab{a}}.

\bibitem[Laskin et~al.(2020{\natexlab{b}})Laskin, Srinivas, and
  Abbeel]{laskin_srinivas2020curl}
Michael Laskin, Aravind Srinivas, and Pieter Abbeel.
\newblock Curl: Contrastive unsupervised representations for reinforcement
  learning.
\newblock \emph{Proceedings of the 37th International Conference on Machine
  Learning, Vienna, Austria, PMLR 119}, 2020{\natexlab{b}}.
\newblock arXiv:2004.04136.

\bibitem[Ma et~al.(2022)Ma, Sodhani, Jayaraman, Bastani, Kumar, and
  Zhang]{ma2022vip}
Yecheng~Jason Ma, Shagun Sodhani, Dinesh Jayaraman, Osbert Bastani, Vikash
  Kumar, and Amy Zhang.
\newblock Vip: Towards universal visual reward and representation via
  value-implicit pre-training.
\newblock \emph{arXiv preprint arXiv:2210.00030}, 2022.

\bibitem[Majumdar et~al.(2023)Majumdar, Yadav, Arnaud, Ma, Chen, Silwal, Jain,
  Berges, Abbeel, Malik, Batra, Lin, Maksymets, Rajeswaran, and
  Meier]{majumdar2023vc1}
Arjun Majumdar, Karmesh Yadav, Sergio Arnaud, Yecheng~Jason Ma, Claire Chen,
  Sneha Silwal, Aryan Jain, Vincent-Pierre Berges, Pieter Abbeel, Jitendra
  Malik, Dhruv Batra, Yixin Lin, Oleksandr Maksymets, Aravind Rajeswaran, and
  Franziska Meier.
\newblock Where are we in the search for an artificial visual cortex for
  embodied intelligence?
\newblock 2023.

\bibitem[Miller et~al.(2021)Miller, Taori, Raghunathan, Sagawa, Koh, Shankar,
  Liang, Carmon, and Schmidt]{Miller2021AccuracyOT}
John Miller, Rohan Taori, Aditi Raghunathan, Shiori Sagawa, Pang~Wei Koh,
  Vaishaal Shankar, Percy Liang, Yair Carmon, and Ludwig Schmidt.
\newblock Accuracy on the line: on the strong correlation between
  out-of-distribution and in-distribution generalization.
\newblock \emph{ArXiv}, abs/2107.04649, 2021.

\bibitem[Nachum et~al.(2019)Nachum, Dai, Kostrikov, Chow, Li, and
  Schuurmans]{nachum2019algaedice}
Ofir Nachum, Bo~Dai, Ilya Kostrikov, Yinlam Chow, Lihong Li, and Dale
  Schuurmans.
\newblock Algaedice: Policy gradient from arbitrary experience.
\newblock \emph{arXiv preprint arXiv:1912.02074}, 2019.

\bibitem[Nair et~al.(2022)Nair, Rajeswaran, Kumar, Finn, and
  Gupta]{nair2022r3m}
Suraj Nair, Aravind Rajeswaran, Vikash Kumar, Chelsea Finn, and Abhinav Gupta.
\newblock R3m: A universal visual representation for robot manipulation.
\newblock In \emph{Conference on Robot Learning (CoRL)}, 2022.

\bibitem[Naseer et~al.(2021)Naseer, Ranasinghe, Khan, Hayat, Khan, and
  Yang]{naseer2021intriguing}
Muzammal Naseer, Kanchana Ranasinghe, Salman Khan, Munawar Hayat, Fahad Khan,
  and Ming-Hsuan Yang.
\newblock Intriguing properties of vision transformers.
\newblock In A.~Beygelzimer, Y.~Dauphin, P.~Liang, and J.~Wortman Vaughan
  (eds.), \emph{Advances in Neural Information Processing Systems}, 2021.
\newblock URL \url{https://openreview.net/forum?id=o2mbl-Hmfgd}.

\bibitem[Oquab et~al.(2023)Oquab, Darcet, Moutakanni, Vo, Szafraniec, Khalidov,
  Fernandez, Haziza, Massa, El-Nouby, Howes, Huang, Xu, Sharma, Li, Galuba,
  Rabbat, Assran, Ballas, Synnaeve, Misra, Jegou, Mairal, Labatut, Joulin, and
  Bojanowski]{oquab2023dinov2}
Maxime Oquab, Timothée Darcet, Theo Moutakanni, Huy~V. Vo, Marc Szafraniec,
  Vasil Khalidov, Pierre Fernandez, Daniel Haziza, Francisco Massa, Alaaeldin
  El-Nouby, Russell Howes, Po-Yao Huang, Hu~Xu, Vasu Sharma, Shang-Wen Li,
  Wojciech Galuba, Mike Rabbat, Mido Assran, Nicolas Ballas, Gabriel Synnaeve,
  Ishan Misra, Herve Jegou, Julien Mairal, Patrick Labatut, Armand Joulin, and
  Piotr Bojanowski.
\newblock Dinov2: Learning robust visual features without supervision, 2023.

\bibitem[Packer et~al.(2018)Packer, Gao, Kos, Kr{\"a}henb{\"u}hl, Koltun, and
  Song]{Packer2018AssessingGI}
Charles Packer, Katelyn Gao, Jernej Kos, Philipp Kr{\"a}henb{\"u}hl, Vladlen
  Koltun, and Dawn~Xiaodong Song.
\newblock Assessing generalization in deep reinforcement learning.
\newblock \emph{ArXiv}, abs/1810.12282, 2018.

\bibitem[Pari et~al.(2022)Pari, Shafiullah, Arunachalam, and
  Pinto]{Pari2022TheSE}
Jyothish Pari, Nur Muhammad~(Mahi) Shafiullah, Sridhar~Pandian Arunachalam, and
  Lerrel Pinto.
\newblock The surprising effectiveness of representation learning for visual
  imitation.
\newblock \emph{ArXiv}, abs/2112.01511, 2022.

\bibitem[Parisi et~al.(2022)Parisi, Rajeswaran, Purushwalkam, and
  Gupta]{Parisi2022TheUE}
Simone Parisi, Aravind Rajeswaran, Senthil Purushwalkam, and Abhinav~Kumar
  Gupta.
\newblock The unsurprising effectiveness of pre-trained vision models for
  control.
\newblock In \emph{International Conference on Machine Learning}, 2022.

\bibitem[Paszke et~al.(2019)Paszke, Gross, Massa, Lerer, Bradbury, Chanan,
  Killeen, Lin, Gimelshein, Antiga, Desmaison, Kopf, Yang, DeVito, Raison,
  Tejani, Chilamkurthy, Steiner, Fang, Bai, and Chintala]{NEURIPS2019_9015}
Adam Paszke, Sam Gross, Francisco Massa, Adam Lerer, James Bradbury, Gregory
  Chanan, Trevor Killeen, Zeming Lin, Natalia Gimelshein, Luca Antiga, Alban
  Desmaison, Andreas Kopf, Edward Yang, Zachary DeVito, Martin Raison, Alykhan
  Tejani, Sasank Chilamkurthy, Benoit Steiner, Lu~Fang, Junjie Bai, and Soumith
  Chintala.
\newblock Pytorch: An imperative style, high-performance deep learning library.
\newblock In \emph{Advances in Neural Information Processing Systems 32}, pp.\
  8024--8035. Curran Associates, Inc., 2019.
\newblock URL
  \url{http://papers.neurips.cc/paper/9015-pytorch-an-imperative-style-high-performance-deep-learning-library.pdf}.

\bibitem[Peng et~al.(2018)Peng, Andrychowicz, Zaremba, and
  Abbeel]{Peng2018SimtoRealTO}
Xue~Bin Peng, Marcin Andrychowicz, Wojciech Zaremba, and P.~Abbeel.
\newblock Sim-to-real transfer of robotic control with dynamics randomization.
\newblock \emph{2018 IEEE International Conference on Robotics and Automation
  (ICRA)}, pp.\  1--8, 2018.

\bibitem[Radford et~al.(2021)Radford, Kim, Hallacy, Ramesh, Goh, Agarwal,
  Sastry, Askell, Mishkin, Clark, Krueger, and Sutskever]{radford2021clip}
Alec Radford, Jong~Wook Kim, Chris Hallacy, Aditya Ramesh, Gabriel Goh,
  Sandhini Agarwal, Girish Sastry, Amanda Askell, Pamela Mishkin, Jack Clark,
  Gretchen Krueger, and Ilya Sutskever.
\newblock Learning transferable visual models from natural language
  supervision.
\newblock \emph{CoRR}, abs/2103.00020, 2021.
\newblock URL \url{https://arxiv.org/abs/2103.00020}.

\bibitem[Radosavovic et~al.(2022)Radosavovic, Xiao, James, Abbeel, Malik, and
  Darrell]{Radosavovic2022}
Ilija Radosavovic, Tete Xiao, Stephen James, Pieter Abbeel, Jitendra Malik, and
  Trevor Darrell.
\newblock Real-world robot learning with masked visual pre-training.
\newblock \emph{CoRL}, 2022.

\bibitem[Raffel et~al.(2020)Raffel, Shazeer, Roberts, Lee, Narang, Matena,
  Zhou, Li, and Liu]{raffel2020exploringthelimits}
Colin Raffel, Noam Shazeer, Adam Roberts, Katherine Lee, Sharan Narang, Michael
  Matena, Yanqi Zhou, Wei Li, and Peter~J. Liu.
\newblock Exploring the limits of transfer learning with a unified text-to-text
  transformer.
\newblock \emph{Journal of Machine Learning Research}, 21\penalty0
  (140):\penalty0 1--67, 2020.
\newblock URL \url{http://jmlr.org/papers/v21/20-074.html}.

\bibitem[Raileanu et~al.(2020)Raileanu, Goldstein, Yarats, Kostrikov, and
  Fergus]{Raileanu2020AutomaticDA}
Roberta Raileanu, Maxwell Goldstein, Denis Yarats, Ilya Kostrikov, and Rob
  Fergus.
\newblock Automatic data augmentation for generalization in deep reinforcement
  learning.
\newblock \emph{ArXiv}, abs/2006.12862, 2020.

\bibitem[Sadeghi \& Levine(2017)Sadeghi and Levine]{Sadeghi2017CAD2RLRS}
Fereshteh Sadeghi and Sergey Levine.
\newblock Cad2rl: Real single-image flight without a single real image.
\newblock \emph{ArXiv}, abs/1611.04201, 2017.

\bibitem[Sax et~al.(2018)Sax, Emi, Zamir, Guibas, Savarese, and
  Malik]{midLevelReps2018}
Alexander Sax, Bradley Emi, Amir~R. Zamir, Leonidas~J. Guibas, Silvio Savarese,
  and Jitendra Malik.
\newblock Mid-level visual representations improve generalization and sample
  efficiency for learning visuomotor policies.
\newblock 2018.

\bibitem[Sermanet et~al.(2018)Sermanet, Lynch, Chebotar, Hsu, Jang, Schaal, and
  Levine]{Sermanet2017TCN}
Pierre Sermanet, Corey Lynch, Yevgen Chebotar, Jasmine Hsu, Eric Jang, Stefan
  Schaal, and Sergey Levine.
\newblock Time-contrastive networks: Self-supervised learning from video.
\newblock \emph{Proceedings of International Conference in Robotics and
  Automation (ICRA)}, 2018.
\newblock URL \url{http://arxiv.org/abs/1704.06888}.

\bibitem[Shan et~al.(2020)Shan, Geng, Shu, and Fouhey]{Shan2020UnderstandingHH}
Dandan Shan, Jiaqi Geng, Michelle Shu, and David~F. Fouhey.
\newblock Understanding human hands in contact at internet scale.
\newblock \emph{2020 IEEE/CVF Conference on Computer Vision and Pattern
  Recognition (CVPR)}, pp.\  9866--9875, 2020.

\bibitem[Shao et~al.(2022)Shao, Shi, Yi, Chen, and Hsieh]{shao2022on}
Rulin Shao, Zhouxing Shi, Jinfeng Yi, Pin-Yu Chen, and Cho-Jui Hsieh.
\newblock On the adversarial robustness of vision transformers.
\newblock \emph{Transactions on Machine Learning Research}, 2022.
\newblock URL \url{https://openreview.net/forum?id=lE7K4n1Esk}.

\bibitem[Tobin et~al.(2017)Tobin, Fong, Ray, Schneider, Zaremba, and
  Abbeel]{Tobin2017DomainRF}
Joshua Tobin, Rachel Fong, Alex Ray, Jonas Schneider, Wojciech Zaremba, and
  P.~Abbeel.
\newblock Domain randomization for transferring deep neural networks from
  simulation to the real world.
\newblock \emph{2017 IEEE/RSJ International Conference on Intelligent Robots
  and Systems (IROS)}, pp.\  23--30, 2017.

\bibitem[Touvron et~al.(2021)Touvron, Cord, Douze, Massa, Sablayrolles, and
  Jegou]{touvron2021deit}
Hugo Touvron, Matthieu Cord, Matthijs Douze, Francisco Massa, Alexandre
  Sablayrolles, and Herve Jegou.
\newblock Training data-efficient image transformers; distillation through
  attention.
\newblock In Marina Meila and Tong Zhang (eds.), \emph{Proceedings of the 38th
  International Conference on Machine Learning}, volume 139 of
  \emph{Proceedings of Machine Learning Research}, pp.\  10347--10357. PMLR,
  18--24 Jul 2021.
\newblock URL \url{https://proceedings.mlr.press/v139/touvron21a.html}.

\bibitem[Weyand et~al.(2020)Weyand, Araujo, Cao, and Sim]{weyand2020GLDv2}
T.~Weyand, A.~Araujo, B.~Cao, and J.~Sim.
\newblock {Google Landmarks Dataset v2 - A Large-Scale Benchmark for
  Instance-Level Recognition and Retrieval}.
\newblock In \emph{Proc. CVPR}, 2020.

\bibitem[Xiao et~al.(2022)Xiao, Radosavovic, Darrell, and Malik]{Xiao2022mvp}
Tete Xiao, Ilija Radosavovic, Trevor Darrell, and Jitendra Malik.
\newblock Masked visual pre-training for motor control.
\newblock \emph{arXiv:2203.06173}, 2022.

\bibitem[Xie* et~al.(2023)Xie*, Lee*, and Finn]{xie2023benchmarking}
Annie Xie*, Lisa Lee*, and Chelsea Finn.
\newblock Benchmarking environment generalization in robotic imitation
  learning, 2023.
\newblock URL \url{https://github.com/RLAgent/factor-envs}.

\bibitem[Xing et~al.(2021)Xing, Gupta, Powers*, and
  Dean*]{xing2021kitchenshift}
Eliot Xing, Abhinav Gupta, Sam Powers*, and Victoria Dean*.
\newblock Kitchenshift: Evaluating zero-shot generalization of imitation-based
  policy learning under domain shifts.
\newblock In \emph{NeurIPS 2021 Workshop on Distribution Shifts: Connecting
  Methods and Applications}, 2021.
\newblock URL \url{https://openreview.net/forum?id=DdglKo8hBq0}.

\bibitem[Yarats et~al.(2021)Yarats, Kostrikov, and Fergus]{yarats2021image}
Denis Yarats, Ilya Kostrikov, and Rob Fergus.
\newblock Image augmentation is all you need: Regularizing deep reinforcement
  learning from pixels.
\newblock In \emph{International Conference on Learning Representations}, 2021.
\newblock URL \url{https://openreview.net/forum?id=GY6-6sTvGaf}.

\bibitem[Yen-Chen et~al.(2020)Yen-Chen, Zeng, Song, Isola, and
  Lin]{lin2020learning}
Lin Yen-Chen, Andy Zeng, Shuran Song, Phillip Isola, and Tsung-Yi Lin.
\newblock Learning to see before learning to act: Visual pre-training for
  manipulation.
\newblock In \emph{IEEE International Conference on Robotics and Automation
  (ICRA)}, 2020.
\newblock URL \url{https://yenchenlin.me/vision2action/}.

\bibitem[Yoneda et~al.(2021)Yoneda, Yang, Walter, and
  Stadie]{yoneda2021invariance}
Takuma Yoneda, Ge~Yang, Matthew~R. Walter, and Bradly Stadie.
\newblock Invariance through latent alignment, 2021.

\bibitem[Yu et~al.(2019)Yu, Quillen, He, Julian, Hausman, Finn, and
  Levine]{yu2019meta}
Tianhe Yu, Deirdre Quillen, Zhanpeng He, Ryan Julian, Karol Hausman, Chelsea
  Finn, and Sergey Levine.
\newblock Meta-world: A benchmark and evaluation for multi-task and meta
  reinforcement learning.
\newblock In \emph{Conference on Robot Learning (CoRL)}, 2019.
\newblock URL \url{https://arxiv.org/abs/1910.10897}.

\bibitem[Zhao et~al.(2019)Zhao, Sigaud, Stulp, and
  Hospedales]{Zhao2019InvestigatingGI}
Chenyang Zhao, Olivier Sigaud, Freek Stulp, and Timothy~M. Hospedales.
\newblock Investigating generalisation in continuous deep reinforcement
  learning.
\newblock \emph{ArXiv}, abs/1902.07015, 2019.

\bibitem[Zhao et~al.(2023)Zhao, Kumar, Levine, and Finn]{zhao2023LearningFG}
Tony Zhao, Vikash Kumar, Sergey Levine, and Chelsea Finn.
\newblock Learning fine-grained bimanual manipulation with low-cost hardware,
  2023.

\end{thebibliography}
\bibliographystyle{iclr2024_conference}

\appendix
\section{Appendix}
\subsection{Pre-Trained Model Details}
\label{sec:model_details}

\textbf{RN-INSUP}~\citep{He2016DeepRL} is a ResNet model trained on the ImageNet classificaiton task. We use the default weights and model provided by the Pytorch~\citep{NEURIPS2019_9015} library.

\textbf{ViT-INSUP} is a Vision Transformer~\citep{dosovitskiy2021an} that has been distilled~\citep{touvron2021deit} from a larger network that was trained on the ImageNet classification task. In our experiments, we use the model weights and architecture provided in \citet{naseer2021intriguing} with a patch size of 16.

\textbf{SIN-SUP} \citep{naseer2021intriguing} trains a vision transformer on Stylized Image-Net (SIN)~\citep{geirhos2018imagenettrained}. The SIN dataset was constructed to increase the degree to which a model makes predictions on shape instead of texture. Our model weights come from \citet{naseer2021intriguing} and we use the non-distilled DeiT~\citep{touvron2021deit} training variant.

\textbf{ViT-DINO}~\citep{caron2021emerging} is trained with extensive augmentations and a self-supervised, contrastive loss that together lead to emergent segmentation within the self-attention heads of the ViT model. We use the model and weights provided by \citet{caron2021emerging}. Interestingly, we don't find the DINO objective to lead to a high shape-bias. This suggests that there are other metrics that measure the degree to which a model is object-centric other than shape-bias.

\textbf{ResNet50-DINO} is learned with the same recipe as ViT-DINO. We use the model and weights from \citet{caron2021emerging}.

\textbf{MoCo. v3, RN} \citep{chen2021mocov3} leverages a contrastive loss with momentum encoding \citep{He2019MomentumCF} of positive targets. It is trained with the same recipe as MoCo. v3, ViT-B.

\textbf{MoCo. v3, ViT-B}~\citep{chen2021mocov3} are trained in a similar manner as the original MoCo \citep{He2019MomentumCF}, but with changes to improve the stability of training, which are specific to the ViT archiecture. We use the checkpoint after 300 epochs.

\textbf{MoCo. v3, ViT-S}~\citep{chen2021mocov3} is trained in a similar manner as MoCo. v3, ViT-B. Even though the smaller model benefits from a longer training horizon, we use the checkpoint at 300 epochs for consistency.

\textbf{MAE-IN, ViT-S} follows the same training recipe as MVP, but on top of the ImageNet dataset. We use the weights provided by \citet{Radosavovic2022}.

\textbf{R3M} \citep{nair2022r3m} trains a ResNet model with a combination of manipulation-specific losses--including a time-contrastive loss \citep{Sermanet2017TCN}, video-language alignemnt loss, and L1-regularization--on the Ego4D \citep{Grauman2022Ego4DAT} dataset.

\textbf{MVP}~\citep{Radosavovic2022} trains a ViT-B for masked autoencoding (MAE) \citep{He2021MaskedAA} on the Ego4D \citep{Grauman2022Ego4DAT}, Something-Something \citep{Goyal2017TheS}, YouTube 100 Days of Hands \citep{Shan2020UnderstandingHH}, EpicKitchens \citep{Damen2018EPICKITCHENS}, and ImageNet \citep{deng2009imagenet} datasets. Unlike R3M, the model is not designed to be exclusive to manipulation.

\textbf{MVP, ViT-S (HOI)}~\citep{Xiao2022mvp} is a predecessor of the model described above that trains a ViT-S/16 with an MAE objective on Something-Something \citep{Goyal2017TheS}, YouTube 100 Days of Hands \citep{Shan2020UnderstandingHH}, EpicKitchens \citep{Damen2018EPICKITCHENS}, and ImageNet \citep{deng2009imagenet}.

\textbf{VIP} \citep{ma2022vip} uses an action-free dual of the Algaedice \citep{nachum2019algaedice} objective to learn representations that are useful for trajectory optimization or reinforcement learning of control tasks unseen during representation pre-training. They train a ResNet-50 on Ego4D with this objective. 

\textbf{CLIP, ViT-B/16}~\citep{radford2021clip} uses contrastive language-image pre-training to learn visual representations trained on an extensive internet datsaet. The learned models exhibit strong zero-shot performance for multiple tasks such as image classification.

\textbf{DiNo v2, ViT}~\citep{oquab2023dinov2} scales~ \citet{caron2021emerging} to more parameters and a larger dataset. The full model is a 1B parameter ViT trained on LVD-142M, which is a 142M frame dataset composed of ImageNet-1k, ImageNet-22k, Google Landmarks~\citep{weyand2020GLDv2}, and a collection of other datasets spanning fine-grained classification, segmentation, depth estimation, and retrieval. The full model is distilled into smaller models. We select the ViT-S distilled model for our experiments. In Table~\ref{tab:pretrained_model_comparison}, we list the augmentations used on the teacher model. The training loop is only lightly modified during distillation. Suprisingly, the v2 model sees worse in- and out-of-domain performance on our evaluation suite in spite of being distilled from a ladrger model trained on a bigger dataset.

\begin{table}[h]
\scriptsize
\centering
\begin{tabular}{|c|c|c|c|c|}
\hline
\textbf{Name} & \textbf{Loss Function} & \textbf{Architecture} & \textbf{Datasets} & \textbf{Augmentations} \\
\hline
\textcolor{sup}{RN-INSUP} & BCE-Loss & ResNet-50 & ImageNet & Random crop, \\
 & &  (23M params) & (1.2M frames) & Horizontal flip \\
\hline
\textcolor{sup}{ViT-INSUP} & BCE-Loss & ViT-S/16 & ImageNet & Random crop, \\
 & & (22M params) & (1.2M frames) & Horizontal flip \\
 \hline
\textcolor{sup}{SIN-SUP} & BCE-Loss & ViT-S/16 & Stylized-ImageNet & Random crop, \\
 & & (22M params) & (1.2M frames) & Horizontal flip \\
 \hline
 \textcolor{selfsup}{ResNet50-DINO} & Distillation & ResNet-50 & ImageNet & Multi-crop, \\
 & & (23M params) & (1.2M frames) & Color-jittering, \\
 & & & & Gaussian blur, \\
 & & & & Solarization \\
\hline
\textcolor{selfsup}{ViT-DINO} & Distillation & ViT-S/16 & ImageNet & Multi-crop, \\
 & & (22M params) & (1.2M frames) & Color-jittering, \\
 & & & & Gaussian blur, \\
 & & & & Solarization \\
% ---------------------------------------------------------------------------------
\hline
 \textcolor{selfsup}{MoCo. v3, RN} & Contrastive & ResNet50 & ImageNet & Resize, \\
 & & (23M params) & (1.2M frames) & Color-jittering, \\
 & & & & Horizontal flip, \\
 & & & & Grayscale, \\
 & & & & Gaussian blur, \\
 & & & & Solarization \\
\hline
 \textcolor{selfsup}{MoCo. v3, ViT-S} & Contrastive & ViT-S/16 & ImageNet & Resize, \\
 & & (22M params) & (1.2M frames) & Color-jittering, \\
 & & & & Horizontal flip, \\
 & & & & Grayscale, \\
 & & & & Gaussian blur, \\
 & & & & Solarization \\
\hline
 \textcolor{selfsup}{MoCo. v3, ViT-B} & Contrastive & ViT-B/16 & ImageNet & Resize, \\
 & & (88M params) & (1.2M frames) & Color-jittering, \\
 & & & & Horizontal flip, \\
 & & & & Grayscale, \\
 & & & & Gaussian blur, \\
 & & & & Solarization \\
\hline
 \textcolor{selfsup}{MAE-IN, ViT-S} & Masked auto-encoding & ViT-S & ImageNet & Random resize, \\
 & & (22M params) & (1.2M frames) & Random crop \\
\hline
\textcolor{manip}{R3M} & Time-contrastive,  & ResNet-50 & Ego4D & Random crop  \\
 & L1-regularization, & (23M params) & (4.3M frames) & \\
 & Video-lang alignment & & & \\ % ---------------------------------------------------------------------------------
\hline
\textcolor{manip}{MVP, ViT-S (HOI)} & Masked auto-encoding & ViT-S & EpicKitchens & None \\
 & & (22M params) & 100 Days of Hands, &  \\
 & & & Something-Something &  \\
 & & & (700k frames) &  \\
\hline
\textcolor{manip}{MVP} & Masked auto-encoding  & ViT-B & Ego4D, ImageNet & None  \\
 & & (88M params) & EpicKitchens, &  \\
 & & & 100 Days of Hands, &  \\
 & & & Something-Something &  \\
 & & & (4.5M frames) &  \\
 \hline
\textcolor{manip}{VIP} &  Algaedice Dual & ResNet-50 & Ego4D & Random crop  \\
 &  & (23M params) & (4.3M frames) &   \\
\hline
\textcolor{other}{CLIP, ViT-B/16} & Contrastive & ViT-B/16 & Internet data & Random crop \\
 & & (88M params) & (400M pairs) & \\
\hline
\textcolor{other}{DiNo v2, ViT} & Distillation & ViT-S/14 & LVD & Multi-crop, \\
 & & (21M params) & (142M frames)  & Color-jittering, \\
 & & & & Grayscale, \\
 & & & & Gaussian blur, \\
 & & & & Solarization \\
\hline
% ---------------------------------------------------------------------------------
\end{tabular}
\caption{List of pre-trained models with corresponding loss function, augmentations, and datasets used for pre-training. We color code by the data and loss type: \textcolor{sup}{ImageNet supervised}, \textcolor{selfsup}{self-supervised}, \textcolor{manip}{trained specifically for manipulation or control tasks}, and \textcolor{other}{other}.}
\label{tab:pretrained_model_comparison}
\end{table}

\subsection{Details of the Environments}
\label{sec:environment_details}

\textbf{FrankaKitchen}~\citep{gupta2019relay} is a simulated kitchen environment with a 9-DoF Franka robot. There a multiple household objects available for interaction. The environment is designed to compose tasks together hierarchically, but we focus on learning policies to successfully complete a single task. The episode length is 50 and we inherit the randomization scheme used in R3M, which randomizes the position of the kitchen at the start of each episode.

\textbf{Meta-World}~\citep{yu2019meta} is a simulated manipulation environment that consists of various table-top manipulation interactions. Unlike FrankaKitchen, the scene objects vary between different tasks. The positions of the objects are randomized at the start of each episode. The maximum episode length is 500.

\subsection{Details of the Disribution Shifts}
\label{sec:dist_shift_details}
\begin{figure*}[t]
    \centering
    \includegraphics[width=\textwidth]{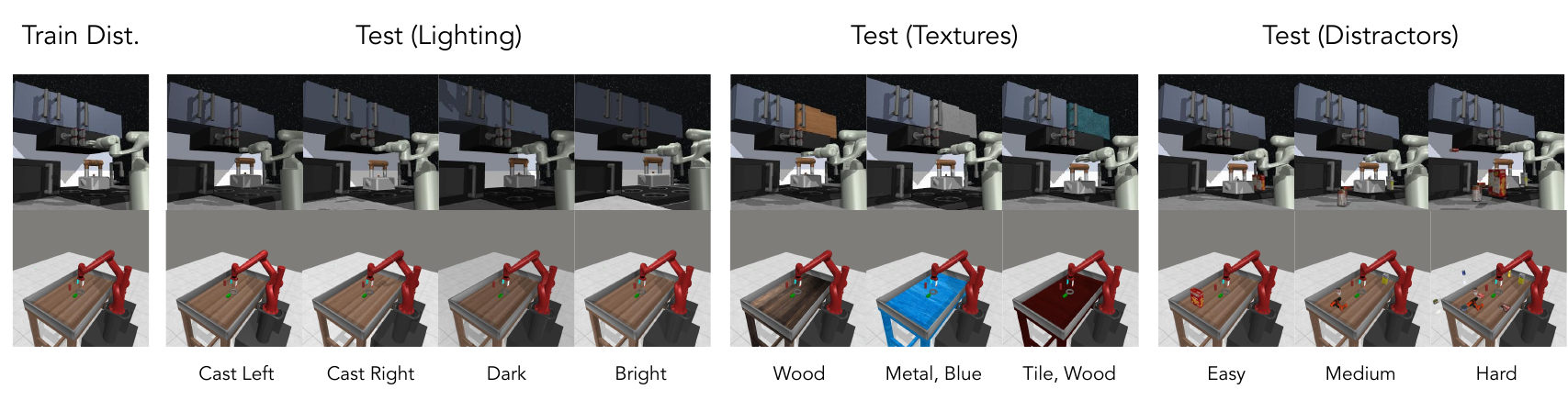}
    \caption{ We visualize each distribution shift from the left camera angle on the FrankaKitchen (top) and Meta-World (bottom) environments.}
    \label{fig:shift_vis}
\end{figure*}

Each distribution shift is visualized from the left camera angle in Figure~\ref{fig:shift_vis}. We don't use the MuJoCo scanned object dataset that is used in \citep{xie2023benchmarking} because of imperfections in the coloring of the textures.

\subsection{Policy Training Details}
\label{sec:bc_details}
\begin{table}[htbp]
  \centering
    \begin{tabular}{cc}
    \toprule
    Hyperparameter & Value \\
    \midrule
    Loss type & MSE \\
    Learning rate & 0.001 \\
    Batch size & 32 \\
    Train steps & 20,000 \\
    Optimizer & Adam \\
    \bottomrule
    \end{tabular}%
  \caption{Hyperparameters for IL Policy Training}
  \label{tab:hyperparameters}
\end{table}

 We learn a 2-layer MLP on top of the pre-trained, frozen features with 10 demonstrations. We use the same expert demonstrations as in R3M. We train policies independently over the `left\_cap2` and `right\_cap2` camera angles and show results averaged over both camera angles. We also provide proprioception to the policy. The final performance is averaged over the task settings for each seed. The hyperparamters for policy training are summarized in Table~\ref{tab:hyperparameters}. Error bars are 95\% confidence interval over seeds.

\begin{figure*}[h!]
    \centering
    \includegraphics[width=.48\textwidth]{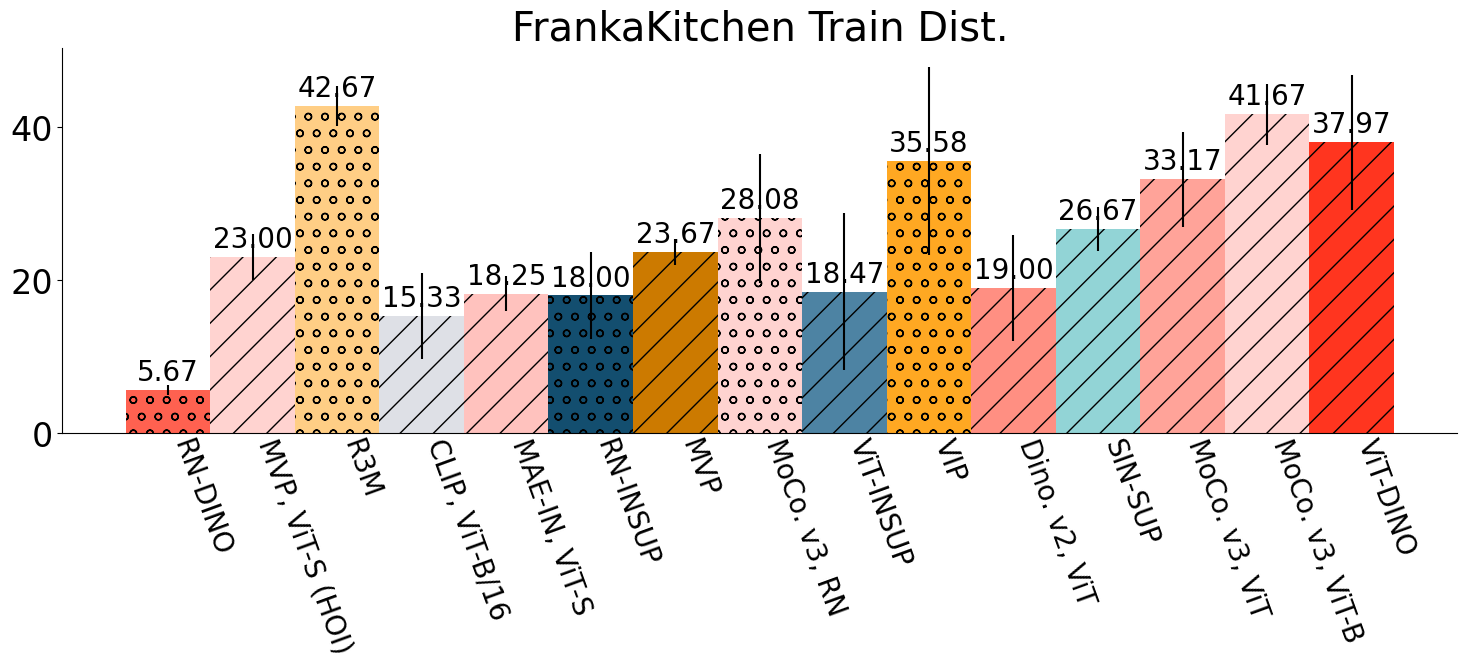}
    \includegraphics[width=.48\textwidth]{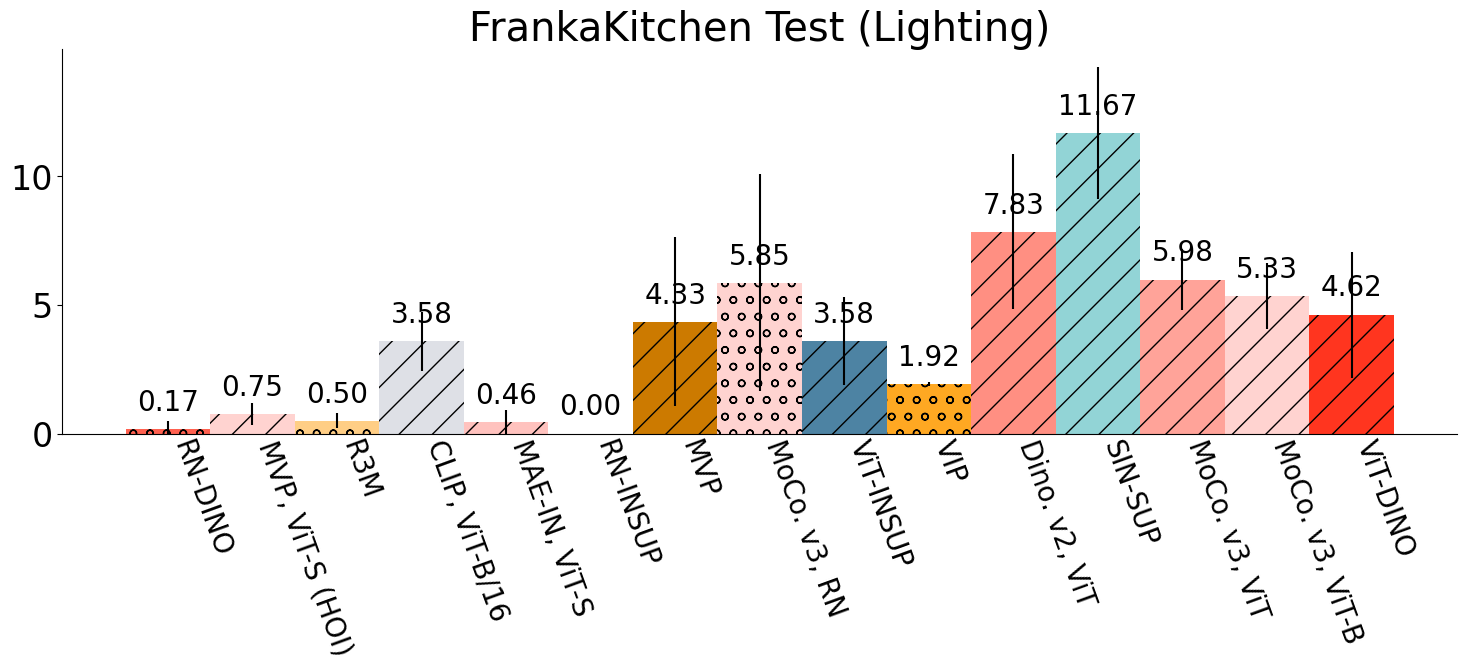}
    \includegraphics[width=.48\textwidth]{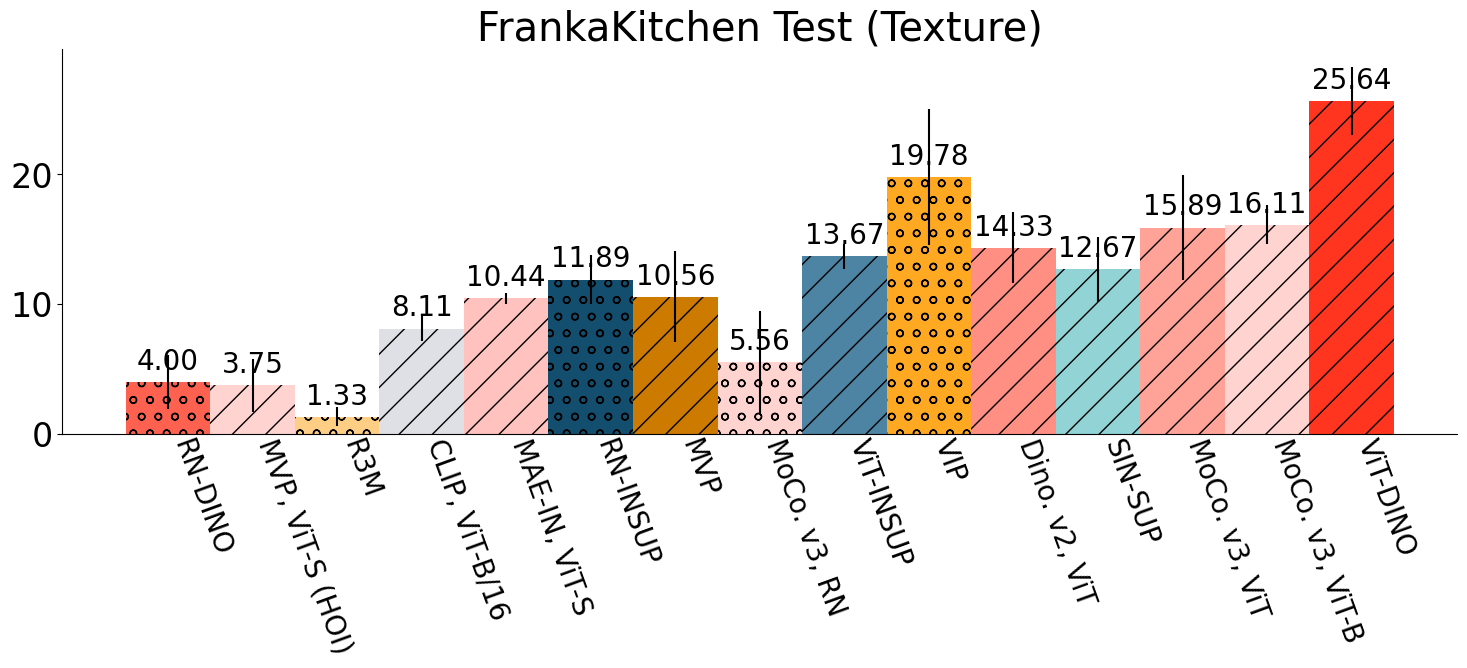}
    \includegraphics[width=.48\textwidth]{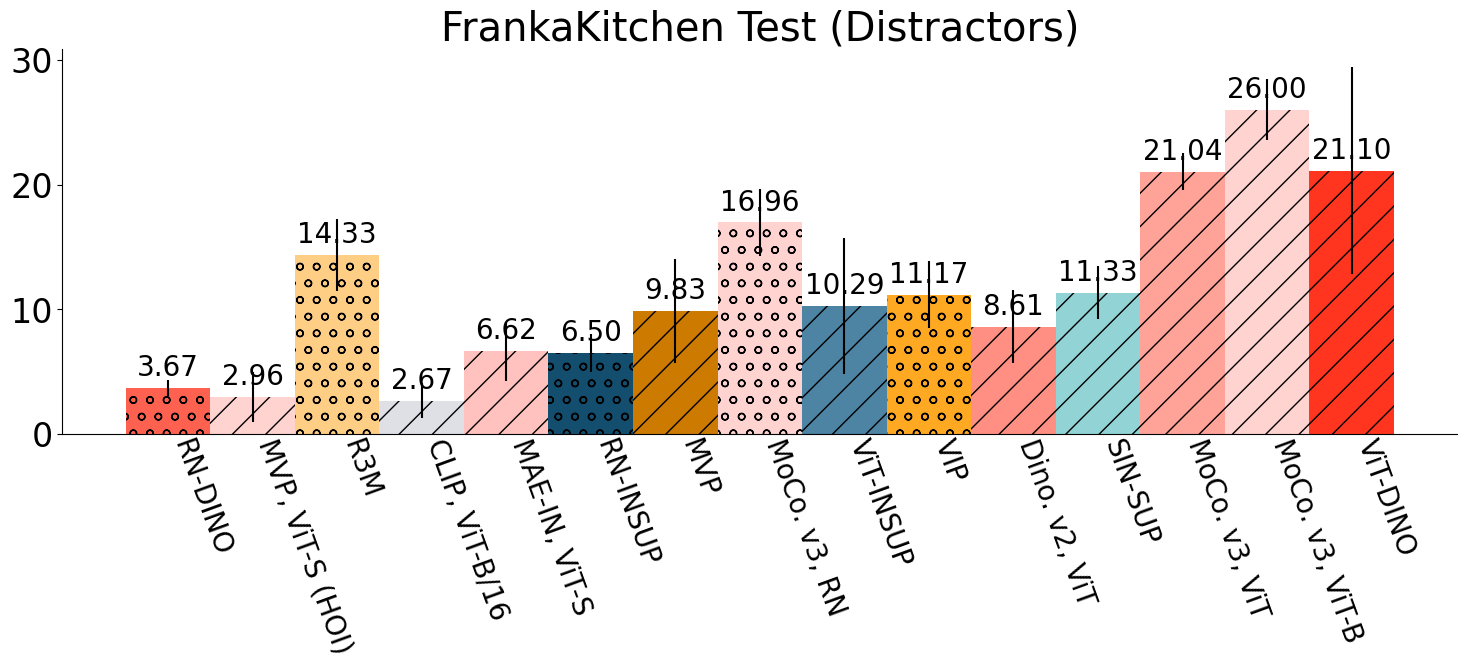}
    \includegraphics[width=.48\textwidth]{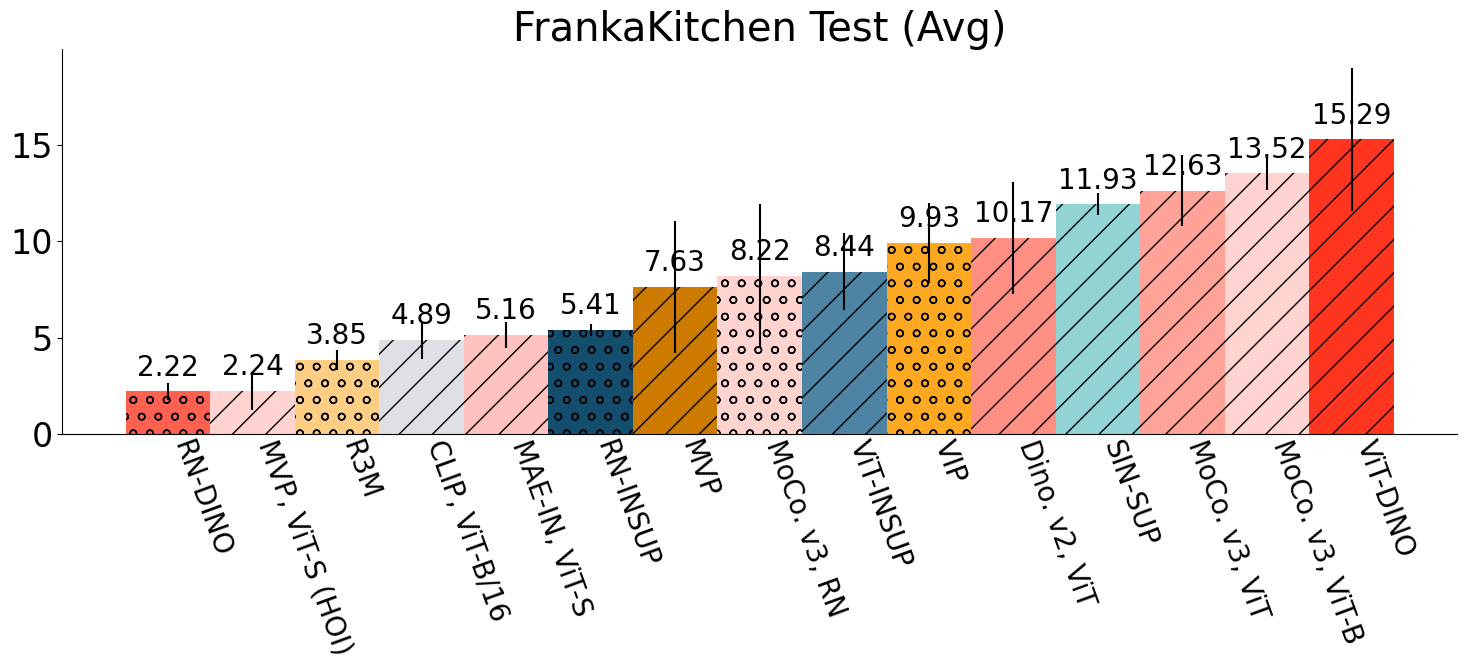}
    \caption{\textbf{Detailed OOD Performance on FrankaKitchen.} 
    }
    \label{fig:ood_kitchen_detailed}
\end{figure*}

\begin{figure*}[h!]
    \centering
    \includegraphics[width=.48\textwidth]{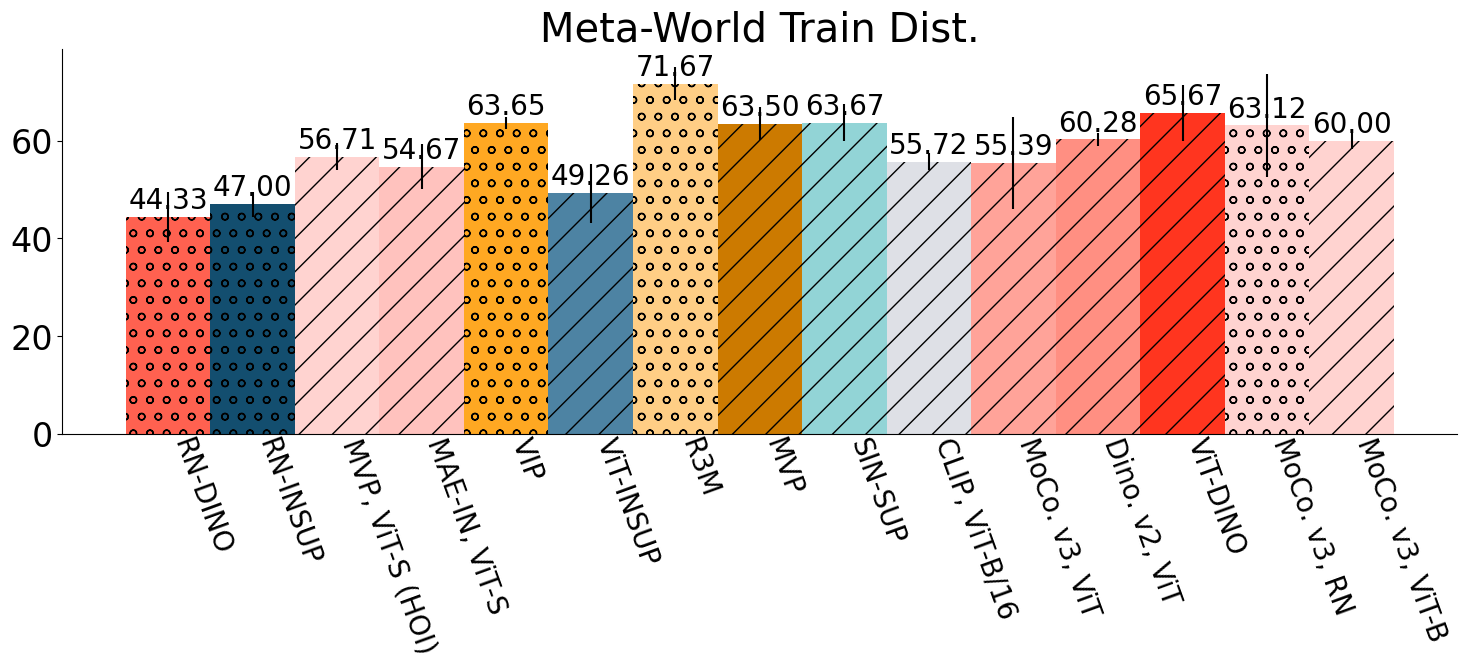}
    \includegraphics[width=.48\textwidth]{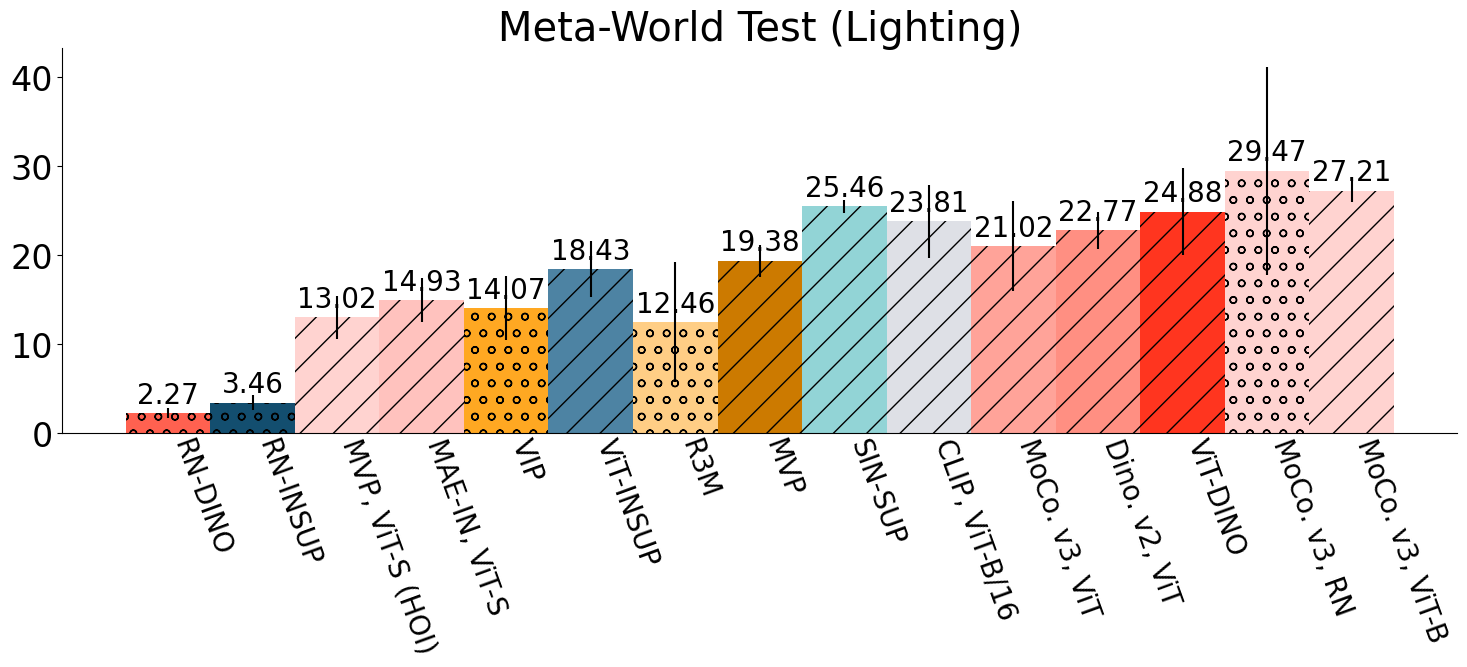}
    \includegraphics[width=.48\textwidth]{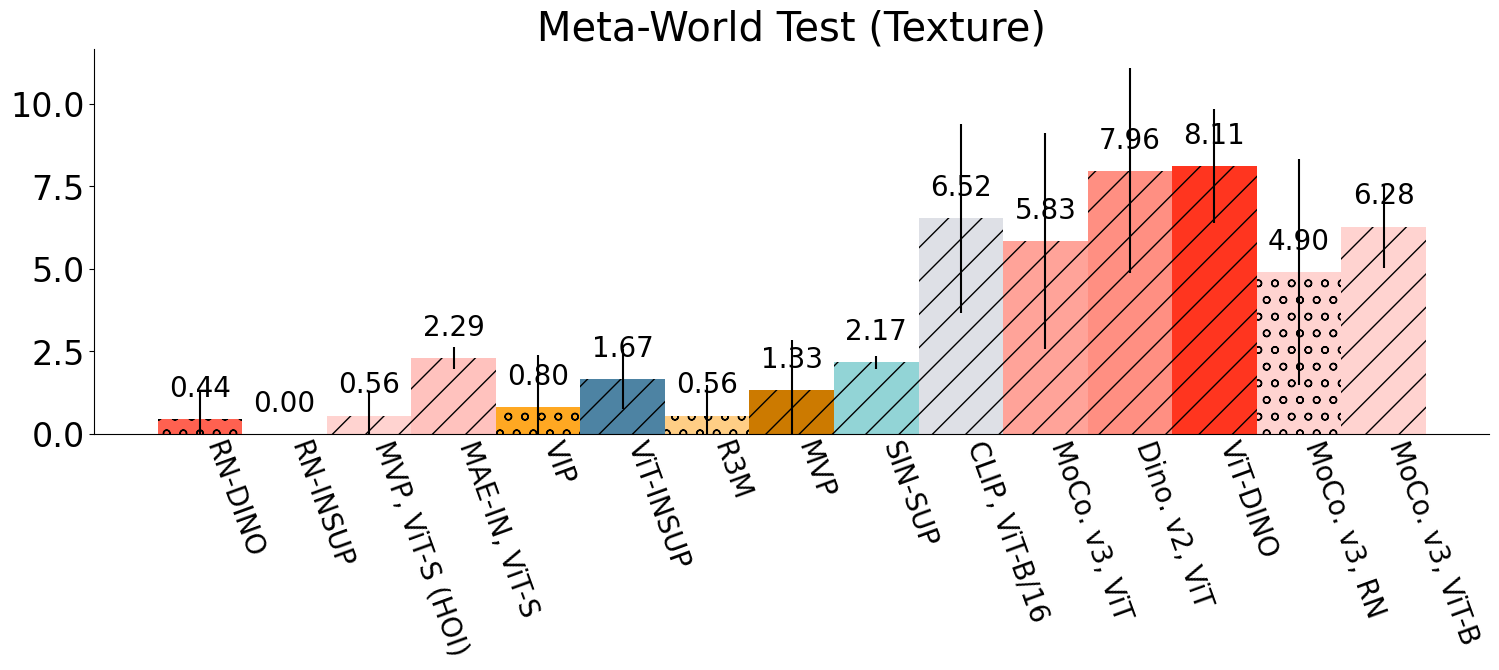}
    \includegraphics[width=.48\textwidth]{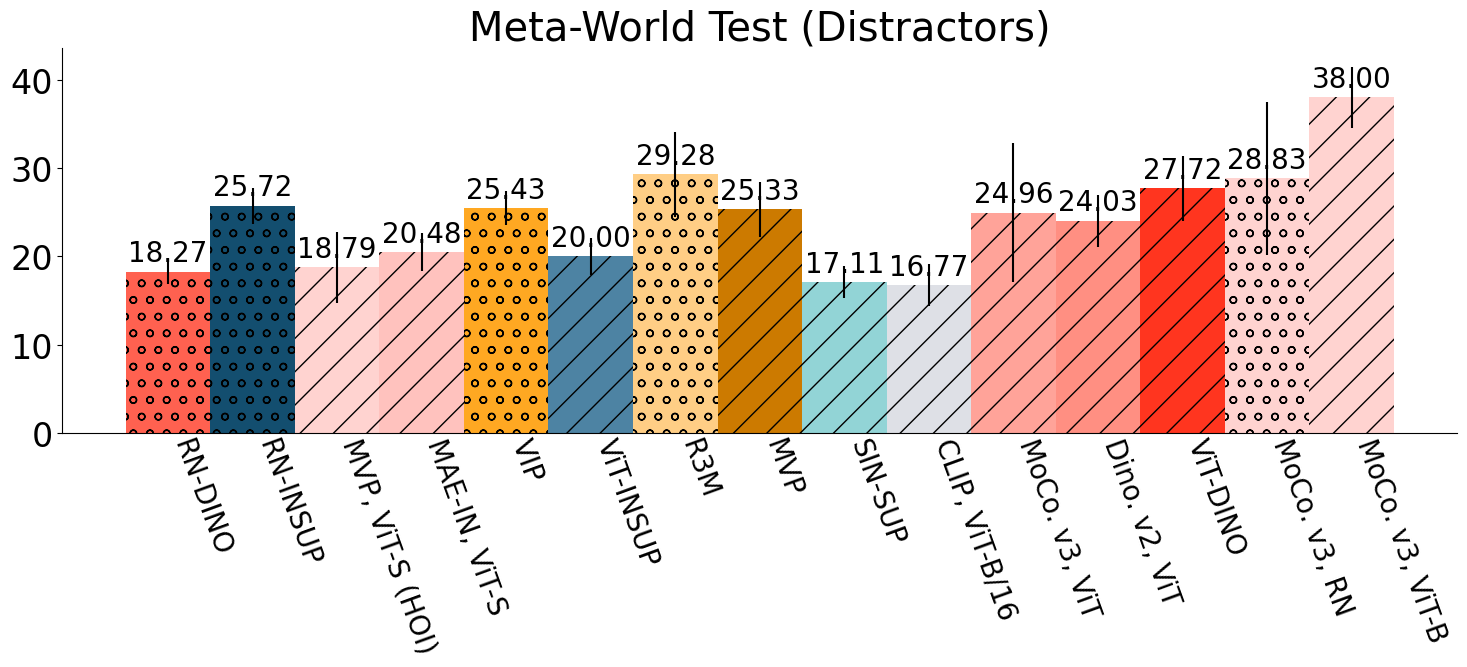}
    \includegraphics[width=.48\textwidth]{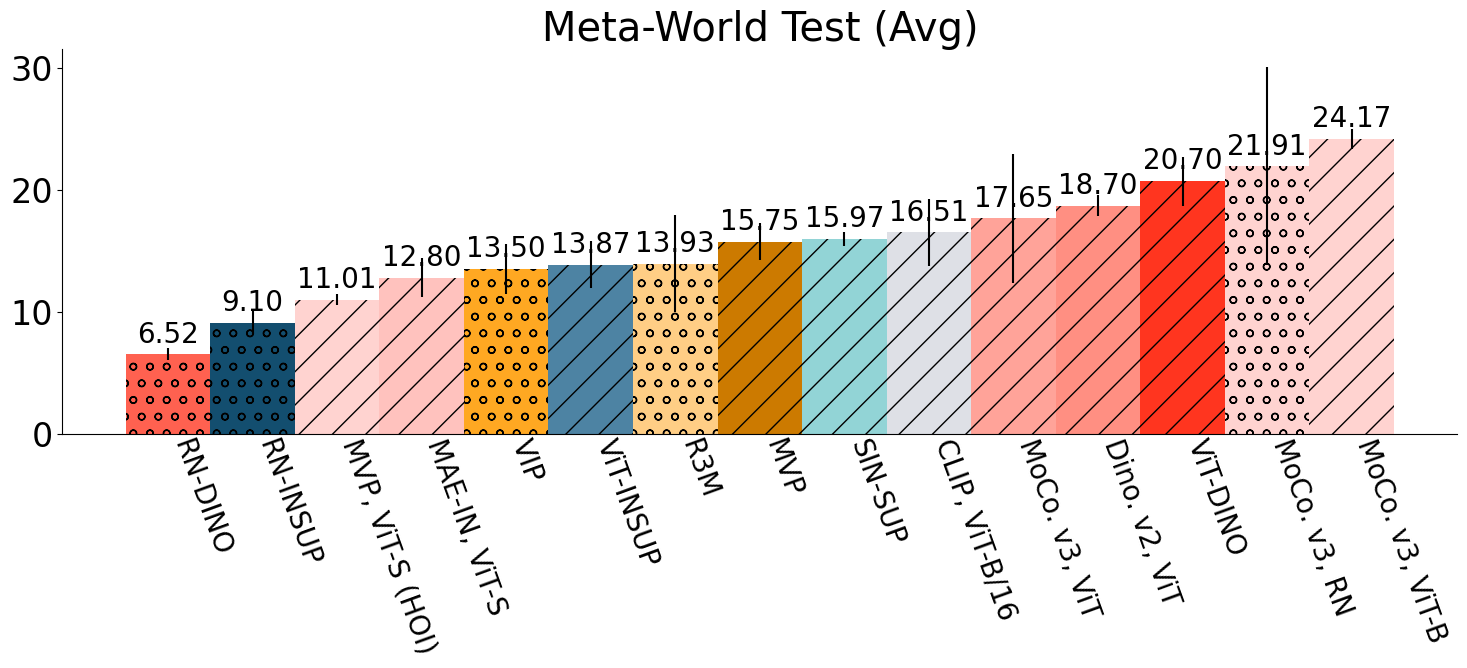}
    \caption{\textbf{Detailed OOD Performance on Meta-World.} 
    }
    \label{fig:ood_metaworld_detailed}
\end{figure*}

\subsection{OOD Perf Details}
\label{sec:ood_perf_details}
To provide a more granular understanding of how the complete set of models performs on our evaluation suite, we break down performance by distribution shift type and environment in Figures~\ref{fig:ood_kitchen_detailed} and \ref{fig:ood_metaworld_detailed}.

\subsection{ImageNet vs OOD Details}
\label{sec:imagenet_vs_ood_details}
To evaluate ImageNet accuracy, we use all publicly available probes that have been trained on top of the frozen model features and evaluate them on the ImageNet validation set. The models with available probes are RN-INSUP, RN-DINO, MoCo. v3 RN, ViT-INSUP, ViT-DINO, MoCo. v3 ViT, Dino v2 ViT, MoCo. v3 ViT, SIN-SUP, and CLIP ViT-B/16 and we use the probes that are provided in the implementations cited in Section~\ref{sec:model_details}.

\subsection{Shape-Bias Details}
\label{sec:shape_bias_details}
We evaluate shape-bias using the `model-vs-human` evaluation framework from \citet{Geirhos2021PartialSI} and use the same probes from Section~\ref{sec:imagenet_vs_ood_details} to get classification results on the SIN validation dataset ($D_{cue-conflict}$).

Notably, \citet{naseer2021intriguing} find that vision transformers are more shape-biased when making classification decisions than equivalently trained convolutional networks. In our results, we don't find vision transformers to be more strongly shape biased. Vision transformers and convolutional networks vary in how they handle spatial resolution: spatial resolution decreases in each layer of ResNet-50 but remains constant within a ViT. This could explain why we see the ViT architecture somewhat obviating the need for shape-bias in our results.

\subsection{Jaccard Index Metric Details}
\label{sec:jaccard_details}
We denote this nonlinear, deterministic transform as $M$. Formally, we compute the Jaccard index by calculating the mIoU on the PASCAL VOC validation set, $D_{Pascal}$:

$$J(x_i,x_j) = \mathbb{E}_{D_{Pascal}}\left[\frac{A\cap B}{A \cup B}\right]$$

Where $A$ is a shorthand for positive classification for the target class by $M(\phi(\cdot))$ and $B$ is a shorthand for positive label for the target class. $J$ is evaluated pixel-wise over image indices $x_i$ and $x_j$.

\subsection{Different Levels of Distractors}
\label{sec:distractors}
\begin{figure*}[h]
    \centering
    \includegraphics[width=\textwidth]{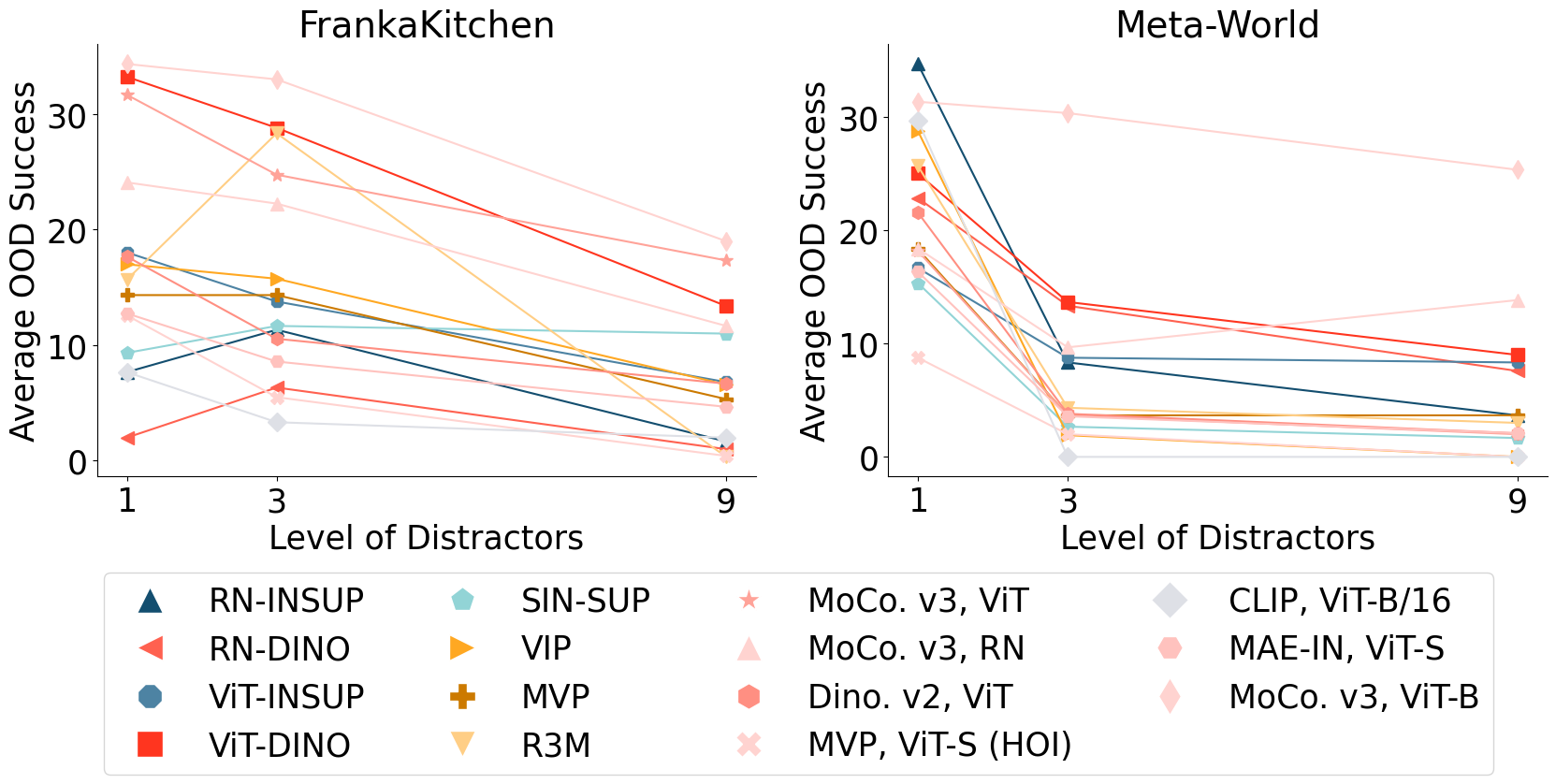}
    \caption{Different levels of distractors.} 
    \label{fig:distractor_levels}
\end{figure*}

We extend Figure~\ref{fig:disrtactors_avg} by including results for ResNets in Figure~\ref{fig:distractor_levels}. Models are color coded using the original color scheme in the paper.

\subsection{Finetuning}
\begin{figure*}[h]
    \centering
    \includegraphics[width=\textwidth]{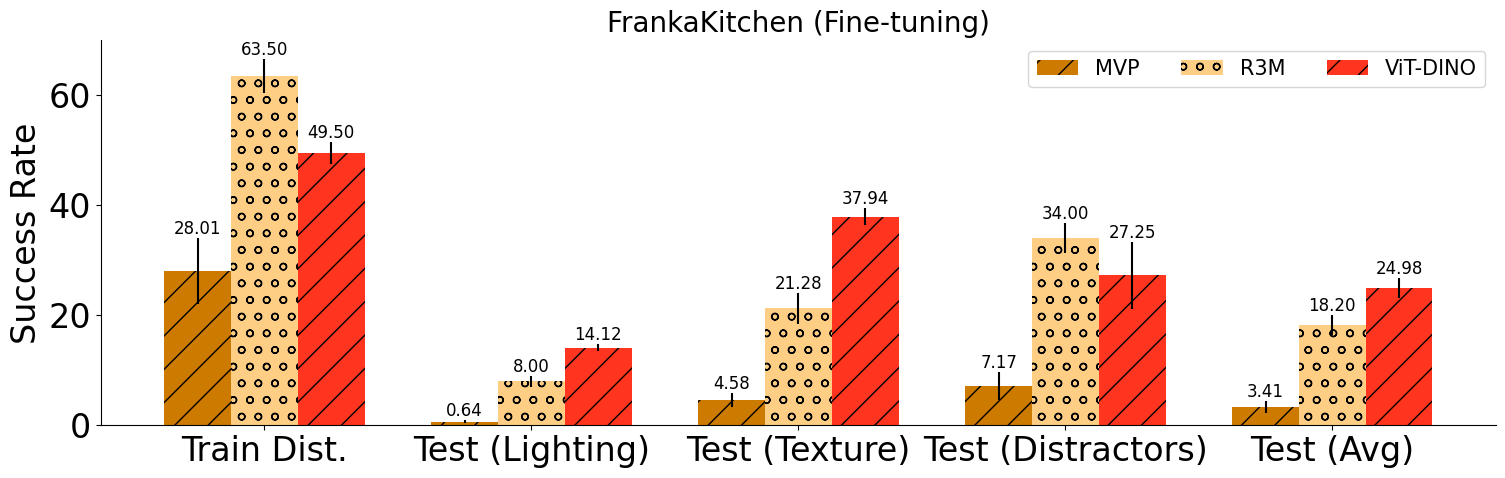}
    \caption{\textbf{Finetuning in FrankaKitchen.} 
    }
    \label{fig:kitchen_ft}
\end{figure*}
Because the goal of this paper is to probe the quality of learned representations, we follow the tradition of performing evaluation on top of frozen model features. This evaluation is also consistent with the increasing view of pre-trained visual representations as ``foundation models'' \citep{bommasani2022opportunities,oquab2023dinov2} that can be deployed without any gradient updates. Nonetheless, even in the fine-tuning regime, in Figure~\ref{fig:kitchen_ft} we still see stronger performance from models that are not designed for manipulation. In this setting, we increased the number of demonstrations to 25 to allow for more data diversity when training the encoders.

% \subsection{Visualizing Representations}

% \begin{figure*}[h!]
%     \centering
%     \includegraphics[width=.48\textwidth]{figures/r3m_comps_scaled.png}
%     \includegraphics[width=.48\textwidth]{figures/dino_dist_comps.png}
%     \caption{We visualize representation distances between each distribution shift.}
%     \label{fig:representation_distance}
% \end{figure*}

\subsection{Real-World Experiment Details}
Our demonstration data contains two subtasks: an initial screwdriver pick-up and then a handover that happen in sequence. We only evaluate success on the subtask of picking up the screwdriver.

\begin{table}[htbp]
  \centering
    \begin{tabular}{cc}
    \toprule
    Hyperparameter & Value \\
    \midrule
    Chunk Size & 100 \\
    KL Weight & 10 \\
    Batch size & 8 \\
    Epochs & 10,000 \\
    Optimizer & Adam \\
    Learning Rate & 1e-5 \\
    \bottomrule
    \end{tabular}%
  \caption{Hyperparameters for Policy Training}
  \label{tab:hyperparameters_real}%
\end{table}

\end{document}